\def\year{2020}\relax
\documentclass[letterpaper]{article} 
\usepackage{aaai20}  
\usepackage{times}  
\usepackage{helvet} 
\usepackage{courier}  
\usepackage[hyphens]{url}  
\usepackage{graphicx} 
\usepackage{graphicx}  
\usepackage{presemble_SL}
\newcommand{\citet}[1]{\citeauthor{#1}\shortcite{#1}}
\newcommand{\citep}{\cite}

\title{Interpreting Adversarial Examples by Activation Promotion and Suppression}

\author{\\Kaidi Xu$^1$, Sijia Liu$^{2}$, Gaoyuan Zhang$^2$, Mengshu Sun$^1$, Pu Zhao$^1$, Quanfu Fan$^2$, Chuang Gan$^2$, Xue Lin$^1$\\
$^1$ Northeastern University\\
$^2$ MIT-IBM Watson AI Lab, IBM Research\\
{\tt\small sijia.liu@ibm.com}
}

\begin{document}

\maketitle

\begin{abstract}
It is widely known that  convolutional neural networks (CNNs) are vulnerable to adversarial  examples: images with imperceptible perturbations crafted to fool classifiers. 
However, interpretability of  these   perturbations is less explored in the literature.
This work aims to better understand the roles of adversarial perturbations and provide visual explanations from pixel, image and network perspectives.
We 
 show that adversaries have a promotion-suppression effect (PSE)  on neurons' activations  and 
can be primarily categorized into three types: 
i) suppression-dominated perturbations that mainly reduce the classification score of the true label, ii) promotion-dominated perturbations that focus on boosting the confidence of the target label, and iii) balanced perturbations that play a dual role in  suppression and promotion. 
 We also provide   image-level interpretability of adversarial examples. This    links PSE of pixel-level perturbations  to  class-specific
 discriminative image regions localized by class activation mapping \citep{zhou2016learning}. Further, we examine the adversarial effect  through network dissection \citep{bau2017network}, which offers concept-level interpretability of hidden units. 
We show that there exists  a tight connection between the units' sensitivity    to adversarial attacks and their interpretability on semantic concepts.
Lastly, we provide some new insights from our interpretation to improve the adversarial robustness of networks. 
\end{abstract}

\section{Introduction}

\textcolor{black}{
Adversarial examples  
are inputs crafted with the intention of fooling machine learning models
\citep{carlini2016hidden,kurakin2016adversarial,athalye2018obfuscated,xu2019topology}.
{Many existing works have} shown that CNNs are vulnerable to adversarial examples with  human imperceptible pixel-level perturbations.
Different types of adversarial attacks were proposed  with a high success rate  of mis-classification. However, understanding these attacks and further interpreting their effects are so far less explored in the literature.
\textcolor{black}{
In this work, we   attempt to study some fundamental  questions as follows:
\textcolor{black}{a) How to interpret the mechanism of   adversarial  perturbations  
at pixel and image levels?   b) Rather than attack generation, how to explain the effectiveness of different adversarial attacks? c) How to explore  the adversarial effects  on the internal response of CNNs? And d) how does the interpretability of adversarial examples help     robustness?}
}
}

\begin{figure}[tb]
   \centering
\hspace*{-0.1in}\begin{tabular}{p{0.17in}p{0.8in}p{0.8in}p{0.8in}}
&  &  \hspace*{-0.1in}
\parbox{0.9in}{\centering \footnotesize CAM}
&  
\hspace*{-0.1in}
\parbox{0.9in}{\centering \footnotesize CAM + \\perturbed pixels}
\\
  \hspace*{0.05in} \rotatebox{90}{\parbox{0.9in}{\centering \footnotesize original image}}
 &   \hspace*{-0.1in} 
\includegraphics[width=0.9in]{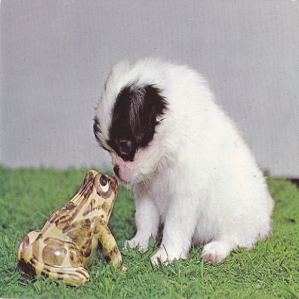}& \hspace*{-0.1in}  
\includegraphics[width=0.9in]{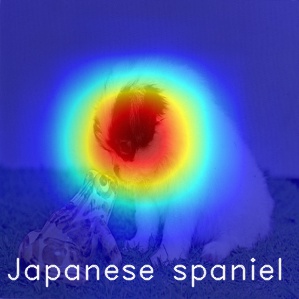}& \hspace*{-0.1in}
\includegraphics[width=0.9in]{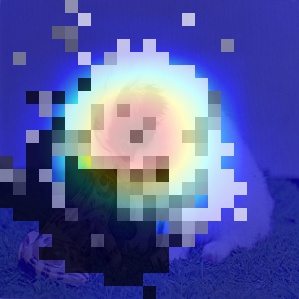}
\\ 
\hspace*{0.05in} \rotatebox{90}{\parbox{0.9in}{\centering \footnotesize adversarial image}}
&   \hspace*{-0.1in}
\includegraphics[width=0.9in]{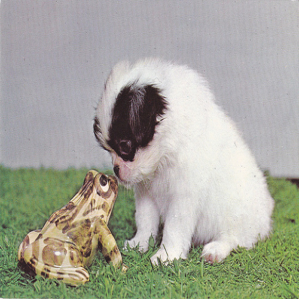}&  \hspace*{-0.1in}
\includegraphics[width=0.9in]{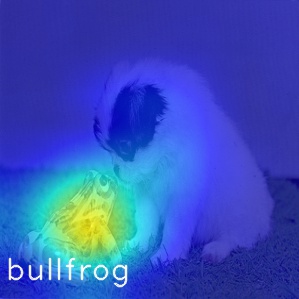}&  \hspace*{-0.1in}
\includegraphics[width=0.9in]{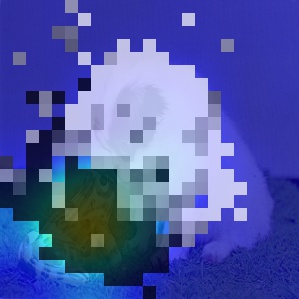}
\end{tabular}
\caption{\footnotesize{Explanation of 
adversarial perturbations  produced by the
C\&W attack \citep{carlini2017towards}. The first column shows the original image (with true label `Japanese spaniel') and its adversarial example (with target label `bullfrog'). The second column demonstrates  CAM of the original image with respect to the true label and  CAM of the adversarial example with respect to the target label. At the third column, the adversarial perturbations are overlaid   on CAM, and their effects  are categorized by our approach: suppression-dominated perturbations (white, at the face of spaniel), promotion-dominated perturbations (black, at the face of bullfrog), and balance-dominated perturbations (gray).
}}
    \label{fig: intro}
\end{figure}


\textbf{Contributions.}
\textcolor{black}{
First, we study the sensitivity and functionality of 
 pixel-level  perturbations on image classification. Unlike  adversarial saliency maps (ASMs)  \citep{papernot2016limitations}, our proposed sensitivity measure  takes into account the dependency among pixels that contribute simultaneously  to the classification confidence. {We uncover the \textit{promotion-suppression   effect (PSE)} of adversarial perturbations.} We group the adversaries  into three types: a) \textit{suppression-dominated perturbations} that mainly reduce the classification score of the true label, b) \textit{promotion-dominated perturbations} that focus on boosting the confidence of the target label, and c)  \textit{balance-dominated  perturbations}  that  play  a  dual  role  in  suppression and promotion. 
}

\textcolor{black}{
{Second, we associate PSE of pixel-level perturbations with image-level  interpretability  based on class  activation map (CAM)  \citep{zhou2016learning}.
We show that the adversarial pattern   can be interpreted using the class-specific discriminative image regions. 
  Figure\,\ref{fig: intro} presents   an   example of the C\&W adversarial attack \citep{carlini2017towards}, 
where suppression- and promotion-dominated perturbations are   matched to the discriminative regions of the natural image (with respect to the true label `Japanese spaniel') and those of the adversarial example (with respect to the target label `bullfrog'), respectively.
We also show that the CAM-based image-level interpretability   provides a means to evaluate the efficacy of  attack generation methods.
Although some works \citep{selvaraju2017grad,xiao2018spatially,xu2018structured} attempted to connect adversarial examples with CAM, they   mainly focused on the visualization of adversarial examples.  
}
}

\textcolor{black}{
Third, 
\textcolor{black}{we present the first attempt to analyze the effect of  adversarial examples on the internal representations of CNNs using the network dissection technique \citep{bau2017network}.
We show a tight connection between the sensitivity of hidden units of CNNs and their interpretability on semantic concepts, which are also aligned with PSE. Furthermore, we  provide some insights on how to improve  robustness by leveraging our interpretation of adversarial examples.
}
}


\textbf{Related Works.}
\textcolor{black}{
The effectiveness of   adversarial attacks \citep{goodfellow2015explaining,KurakinGB2016adversarial,carlini2017towards,chen2017ead,xu2018structured,ye2019second} 
are commonly measured from   attack success rate as well as  $\ell_p$-norm distortion between natural  and  adversarial examples. Some works \citep{karmon2018lavan,brown2017adversarial} generated adversarial attacks by   adding unconstrained noise patches, which \textcolor{black}{are} different from norm-ball constrained attacks, leading to higher noise visibility. 
Rather than attack generation,  
the goal of this paper 
is to understand and explain the effect of 
imperceptible perturbations. Here we focus on norm-ball constrained  adversarial attacks.
Many defense methods have also  been developed against adversarial attacks. Examples include defensive distillation \citep{papernot2016distillation}, random mask \citep{luo2018random}, training with a Lipschitz regularized loss function \citep{finlay2018improved}, and robust adversarial training using min-max optimization   \citep{madry2017towards,sinha2018certifying}. 
}

\textcolor{black}{Although the study on attack generation and defense  has attracted an increasing amount of attention,  interpretability of these examples is less explored in the literature. 
\textcolor{black}{Some preliminary works \citep{papernot2016limitations,yu2018asp} were made on 
evaluating the impact of 
pixel-level adversarial perturbations on changing the classification results.
In \citep{papernot2016limitations}, Jacobian-based ASM 
{was introduced to greedily perturb pixels} that significantly contribute to the likelihood of target
classification. However, ASM 
implicitly ignores the coupling effect of pixel-level perturbations, and it becomes less effective when
 an image has multiple color channels  given the fact that each color channel is treated independently.}
 As an extension of \citep{papernot2016limitations}, the work \citep{yu2018asp} proposed an  adversarial saliency prediction (ASP)  method, which characterizes the divergence of the ASM distribution and the  distribution of perturbations.
}

 \textcolor{black}{
 Both ASM \citep{papernot2016limitations} and ASP \citep{yu2018asp} have helped   humans to understand how adversarial perturbations   made to inputs will affect the outputs of neural networks, however, it remains difficult to visually explain the mechanism of the adversary given the fact that   pixel-level perturbations are small and imperceptible to humans. 
The work  \citep{selvaraju2017grad,xiao2018spatially} adopted CAM to visualize the change of attention regions of the natural and adversarial images, but the use of CAM is   preliminary and its connection with interpretability of pixel-level  perturbations  is  missing. The  most relevant work to ours is \citep{xu2018structured}, which proposed an interpretability score via ASM and CAM. However, it focuses on generating structure-driven adversarial attacks by promoting  group sparsity of perturbations. In contrast, 
we provide more thorough and insightful quantitative    analysis. In particular, we
associate the class-specific discriminative image regions with pixel-level perturbations. 
We also show that the CAM-based   interpretability 
provides means to examine the     effectiveness of   perturbation patterns.
 }
 
 \textcolor{black}{From the network perspective, the work
  \citep{dong2017towards} investigated 
  the effect of an ensemble  attack on neurons' activations.
\textcolor{black}{In \citep{carter2019activation}, Activation Atlas was proposed to show  feature visualizations of   basis neurons as well as common combinations of neurons. And it was applied to visualizing the effect of adversarial patches (rather than norm-ball constrained adversarial perturbations).} 
Different from \citep{carter2019activation,dong2017towards}, we adopt the technique of network dissection \citep{bau2017network,bau2018visualizing} to peer into the effect of adversarial examples on the 
concept-level interpretability  of hidden units. 
}
 


\section{Preliminaries: Attack, Dataset, and Model}
{
Let $\mathbf x_0 \in \mathbb R^n$ denote the \textit{natural} image, and $\boldsymbol{\delta}$ be \textit{adversarial perturbation}   to be designed. Here, unless specified otherwise,  the vector representation of an image is used. The \textit{adversarial example} is then given by $\mathbf x^\prime = \mathbf x_0 + \boldsymbol{\delta}$. By setting the input of the CNNs as $\mathbf x_0$ and $\mathbf x^\prime$, the classifier will predict   the true label $t_0$ and the target label $t$ ($ \neq t_0$), respectively. 
To find the minimum adversarial perturbation  $\boldsymbol{\delta}$ for misclassification from $t_0$ to $t$,
a so-called norm-ball constrained attack technique is commonly used; Examples considered in this paper include \textbf{IFGSM} \citep{goodfellow2015explaining}, \textbf{C\&W} \citep{carlini2017adversarial}, \textbf{EAD} \citep{chen2017ead}, and \textbf{Str} attacks  \citep{xu2018structured}.  
\textcolor{black}{We refer readers to Appendix\,\ref{app: attack_generation} for more details on attack generation.}

}

\textcolor{black}{Our work attempts  to interpret adversarial examples from the pixel   (Sec.\,\ref{sec: pixel}), image   (Sec.\,\ref{sec: image}) 
and network  (Sec.\,\ref{sec: network}) perspective. At  pixel and image levels, we generate adversarial examples from ImageNet under network models Resnet\_v2\_101 \citep{he2016deep} and Inception\_v3 \citep{Szegedy2016RethinkingTI}.
At the network level,  we generate adversarial examples from the 
 Broadly and Densely Labeled Dataset (Broden)  \citep{bau2017network},  {which contains examples with pixel-level concept annotations related to multiple concept categories including} color, material, texture, part, scene  and object. 
 The considered network model is Resnet\_152 \citep{he2016deep}. 
}

\section{Effects of Pixel-level  Perturbations}
\label{sec: pixel}

\textcolor{black}{
We begin by 
quantifying
how much impact a  perturbation 
could make on prediction confidence. 
We use the change of logit scores with respect to (w.r.t.)    both correct and target labels to measure such an effect. This is in the similar spirit of  the C\&W attack loss \citep{carlini2017adversarial} but we focus on grid region-level perturbations as well as the leave-one-out interpretability criterion \citep{yang2019ml}. 
We  build a pixel-level sensitivity measure that  can be further used for image-level sensitivity analysis. We aim to answer the  questions: How to tag the  role of    pixel-level perturbations? And how is the  perturbation sensitivity  associated with the perturbation strength?
}

\textcolor{black}{
Recall that $\mathbf x = \mathbf x_0$ denotes the natural image, and $\mathbf x = \mathbf x^\prime $ corresponds to the adversarial example. We divide an image $\mathbf x$ into $m$ \textit{grid regions} with coordinate sets $\{ \mathcal G_i \}_{i=1}^m$, where each $\mathcal G_i$ contains a group of pixels, and
$\cup_{i=1}^m \mathcal G_i = [n]$. 
Here $[n]$ denotes the overall set of pixels $\{ 1,2,\ldots, n\}$. 
We note that  a proper size of a grid region facilities 
visual explanation   on   semantic image sub-regions and save computation cost. So we set it as $13 \times 13 $ for ImageNet empirically and also consistent with \citep{xu2018structured}.  
Let $\boldsymbol{\delta}_{\mathcal G_i} \in \mathbb R^n$ be
the perturbation at the grid region $\mathcal G_i$, where $[\boldsymbol{\delta}_{\mathcal G_i}]_j = \delta_j$ if $j \in \mathcal G_i$, and $0$ otherwise. Here $[\mathbf a]_i$ or $a_i$ denotes the $i$th element of  $\mathbf a$. We next propose a sensitivity measure of $\boldsymbol{\delta}_{\mathcal G_i} $, which characterizes the impact of pixel-level perturbations on 
  the prediction confidence (in terms of   logit score).
}

\begin{mydef}\label{eq: def_d0t} 
(Sensitivity measure of perturbations): The impact of    perturbation $\boldsymbol{\delta}_{\mathcal G_i}$   on prediction is measured from two aspects: i) the logit change $d_{0,i}$ w.r.t. the true label $t_0$, and ii) the logit change $d_{t,i}$ w.r.t. the target label $t$. That is,
{\small \begin{align}
    &d_{0,i} = \max \{  Z(\mathbf x^\prime - \boldsymbol{\delta}_{\mathcal G_i})_{t_0} - Z(\mathbf x^\prime )_{t_0} , \xi \}, \label{eq: d0i}\\
    & d_{t,i} =  \max \{  Z(\mathbf x^\prime )_{t} - Z(\mathbf x^\prime - \boldsymbol{\delta}_{\mathcal G_i})_{t}, \xi \}, \label{eq: dti} \\
    & s_i = d_{0,i} + d_{t,i} \label{eq: s_d}
\end{align}}%
for $i \in [m]$, where   $Z(\mathbf x)_j$ gives the logit score with respect to class  $j$, and $\xi >0$ is a small positive number.
\end{mydef}

The rationale behind Definition\,\ref{eq: def_d0t} is that the adversarial example $\mathbf x^\prime$, when it is successfully generated, makes the network misclassified from $t_0$ to $t$. Thus, we assign each $\boldsymbol{\delta}_{\mathcal G_i}$  i) the change in the confidence of the true label $t_0$ and ii) the change in the confidence of the target label $t$
when  $\boldsymbol{\delta}_{\mathcal G_i}$  is \textit{removed} from the adversarial example $\mathbf x^\prime$.
The sensitivity measure in Definition\,\ref{eq: def_d0t}  enjoys the similar spirit of   
leave-one-out interpretability   \citep{zeiler2014visualizing,yang2019ml}, but the latter focuses on the prediction sensitivity of a single class    rather than both $t_0$ and $t$.  

\begin{figure}[tb]  
\centering
\vspace*{-4mm}\includegraphics[width=0.47\textwidth]{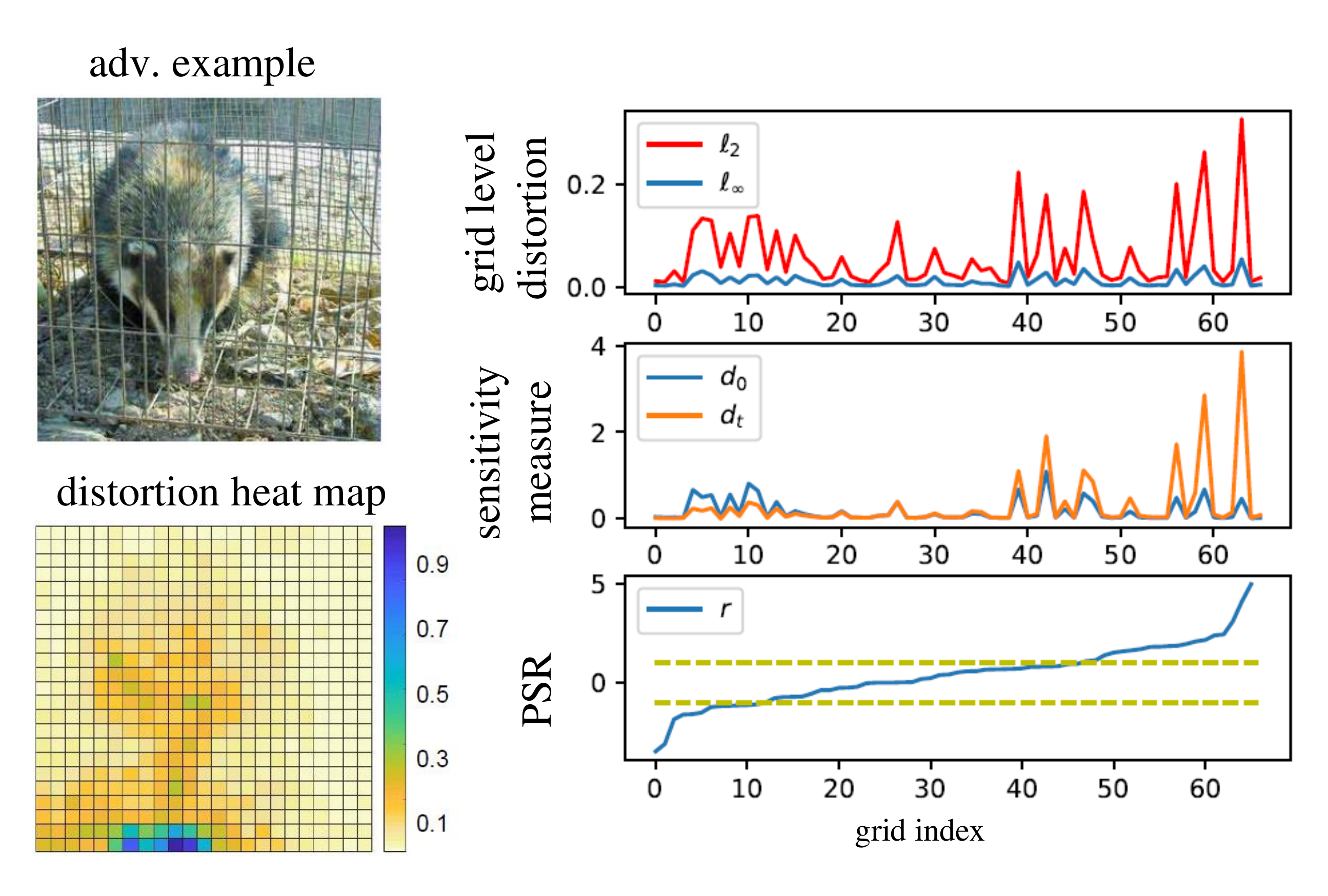}
\caption {\footnotesize{Illustration on sensitivity measure 
via the `badger'-to-`computer' adversarial example generated by C\&W attack. Here the true label is `badger' and the target label is `computer'.
\textcolor{black}{The first column shows the adversarial example and the heat map of $\ell_2$-norm distortion at each grid region, i.e.,  $\{ \| \boldsymbol{\delta}_{\mathcal G_i} \|_2 \}$. 
The second column presents $\ell_p$  norm of $\boldsymbol{\delta}_{\mathcal G_i}$ ($p = 2,\infty$), sensitivity scores $  d_{0,i}  $ and $   d_{t,i}  $, and   PSR $r_i$ versus the index of grid regions, where  the dash lines correspond to the PSR threshold $\pm 1$. 
} 
}
}
\label{fig: d0dtexample}
\end{figure}

In Definition\,\ref{eq: def_d0t}, 
a large $d_{0,i}$ implies a more significant role of a perturbation $\boldsymbol{\delta}_{\mathcal G_i}$ on \textit{suppressing} the classification result away from $t_0$. By contrast, $d_{t,i}$
measures the effect of $\boldsymbol{\delta}_{\mathcal G_i}$ on \textit{promoting} the prediction confidence of the target label. The overall adversarial significance $s_i$  is the combined effect of $d_{0,i}$ and $d_{t,i}$.
Thus, pixels with   small values  of $s_i$ play   less significant roles in misleading  classifiers.
In  \eqref{eq: d0i}--\eqref{eq: dti}, we use $\xi$ to get rid of the negative values of $d_{0,i}$ and $d_{t,i}$, i.e., the  insignificant cases for the adversary.

{With the aid of $\{ d_{0,i} \}$ and $\{ d_{t,i} \}$ in  Definition\,\ref{eq: def_d0t}, we    define a promotion-suppression  ratio (PSR) 
{\small \begin{align}\label{eq: ri}
    r_i = \log_2 \left ( {d_{t,i}}/{d_{0,i}} \right ), i \in [m],
\end{align}}%
which describes the mechanism of  $\boldsymbol{\delta}_{\mathcal G_i}$
on misclassification. \textcolor{black}{In \eqref{eq: ri}, the logarithm is taken for ease of studying PSR under different regimes, e.g, $ r_i \geq 1$ implies that $d_{t,i} \geq 2 d_{0,i}$. Here we   categorize}
the effect of  $\boldsymbol{\delta}_{\mathcal G_i}$ into three types. 
If $r_i < -1$, then we call $\boldsymbol{\delta}_{\mathcal G_i}$   a
\textit{suppression-dominated perturbation}, which is mainly used to   reduce the classification logit of the true  label. 
 If $r_i > 1$, then  we call $\boldsymbol{\delta}_{\mathcal G_i}$ a  
 \textit{promotion-dominated perturbation}, which is mainly used to   boost the classification logit of the target label. If $r_i \in [-1, 1]$, then we call $\boldsymbol{\delta}_{\mathcal G_i}$  a
\textit{balance-dominated perturbation} that  plays  a  dual  role  in  suppression and promotion. Although 
different threshold values on $r_i$ can be used,  we choose $\pm 1$  for  ease of visual explanation.  In Figure \ref{fig: d0dtexample}
we illustrate the sensitivity measures \eqref{eq: d0i}-\eqref{eq: ri}
through an adversarial example   generated by the C\&W attack \citep{carlini2017towards}.
As we can see, either $ d_{0,i} $ or $ d_{t,i} $ (an thus $ s_i $) is correlated with the perturbation strength   at each grid region (in terms of the $\ell_p$  norm of $\boldsymbol{\delta}_{\mathcal G_i}$). 
Also, PSR implies that
most of perturbations in this example contribute to \textcolor{black}{promoting}
the prediction confidence of the target class. This will also be verified by the image-level interpretability  in   Figure\,\ref{fig: CAM_examples_split}.  
}

\begin{figure}[!t]
\centering
\begin{tabular}{cc}
\hspace*{-0.13in} \includegraphics[width=1.6in]{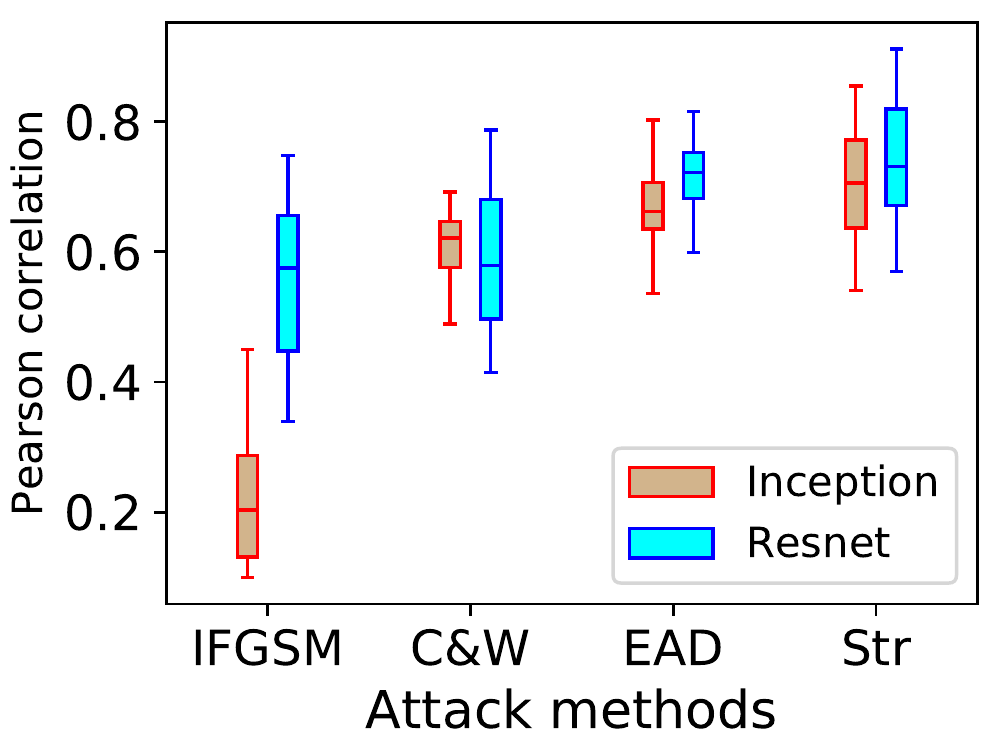} &
\hspace{-1mm}\includegraphics[width=1.6in]{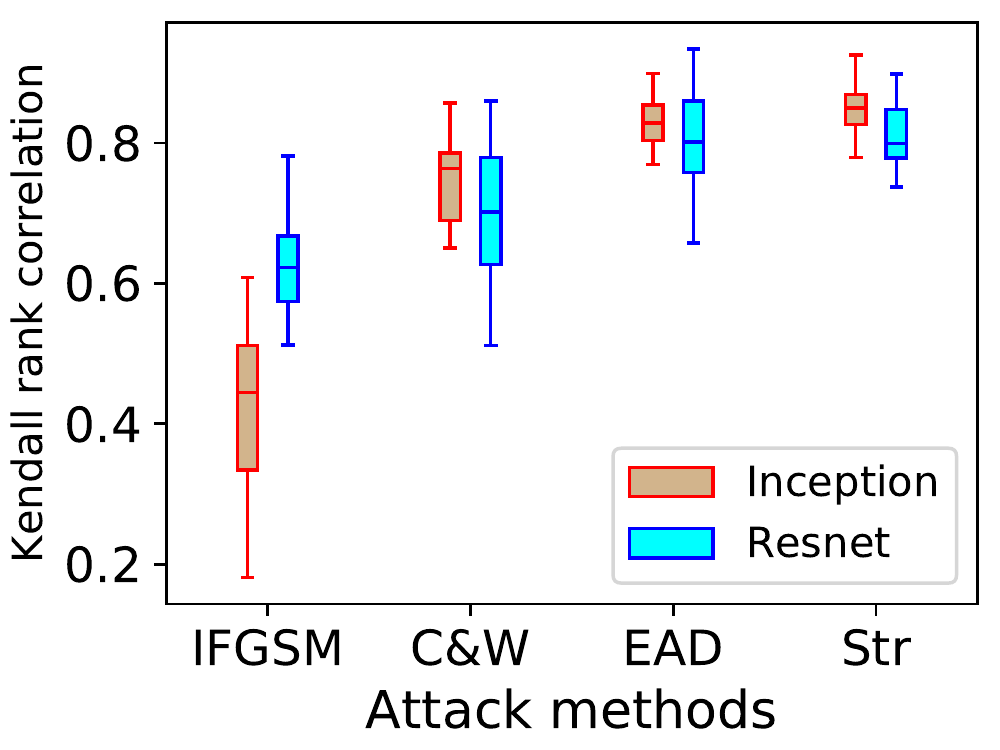}
\end{tabular}
\vspace*{-0.1in}
\caption{Correlation between sensitivity scores   $\{ s_i \}$  and  $\ell_2$  distortion values $\{ \| \boldsymbol{\delta}_{\mathcal G_i} \|_2 \}$. Left: Pearson correlation. Right:   Kendall rank correlation. 
}
\label{figure:correlation}
\end{figure}

\textcolor{black}{The   example of the C\&W attack in Figure\,\ref{fig: d0dtexample} suggests that  the strength of pixel-level perturbations (e.g., in terms of $\ell_2$ norm)  might be strongly correlated with   the    input sensitivity scores.
For a more thorough   quantitative analysis, we examine $5000$  adversarial examples generated by $4$ attack methods under $2$ network models from ImageNet. 
\textcolor{black}{In Figure\,\ref{figure:correlation}, we present Pearson correlation and Kendall rank correlation between the distortion strength $\{ \| \boldsymbol{\delta}_{\mathcal G_i} \|_2 \}$ and the proposed sensitivity scores $\{ s_i \}$ given by \eqref{eq: s_d}. We see that   C\&W, EAD and Str   attacks exhibit a  relatively stronger correlation than IFGSM since the latter perturbs every pixel due to the use of sign operation, while the former attacks   are generated by $\ell_1$ and $\ell_2$-norm penalized optimization  methods that often yield non-uniform and sparser perturbations. For example, the sparest Str-attack   has the highest correlation since it   perturbs   discriminative image regions of high sensitivity \citep{xu2018structured}.  In the next section, we will  further connect pixel-level perturbations   with   discriminative image regions. 
}
}

\section{Interpreting Adversarial Perturbations via Class Activation Map (CAM)}
\label{sec: image}

CAM \citep{zhou2016learning} and other  visual explanation techniques such as GradCAM  \citep{selvaraju2017grad},  GradCAM++ \citep{chattopadhay2018grad} and RISE 
\citep{petsiuk2018rise}
build  a  localizable deep representation, which
exposes the implicit attention of CNNs on a labelled image \citep{zhou2016learning}. In this section, we   analyze the promotion-suppression effect (PSE) of 
adversarial perturbations via image-level interpretability.
We restrict our analysis   to CAM, but     can readily be  extended to other interpretability methods. 

\begin{figure}[t]
   \centering
\begin{tabular}{p{0.45in}p{0.45in}p{0.45in}p{0.45in}p{0.45in}}
  \parbox{0.55in}{   \centering \footnotesize{$F(\bm x^\prime , t_0)$} }
 &  \parbox{0.55in}{  \centering \footnotesize{$F(\bm x_0 , t_0)$} }
 & \parbox{0.55in}{  \centering \footnotesize{$\mathbf x_0$ } }
 & \parbox{0.55in}{  \centering \footnotesize{$F(\bm x_0 , t)$} }
 & \parbox{0.55in}{ \centering \footnotesize{ $F(\bm x^\prime , t)$} }
 \\[-1pt]
\includegraphics[width=0.55in]{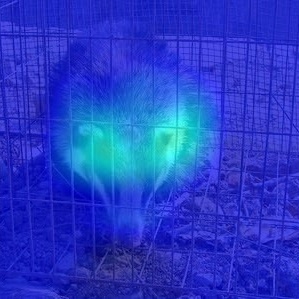}&
\includegraphics[width=0.55in]{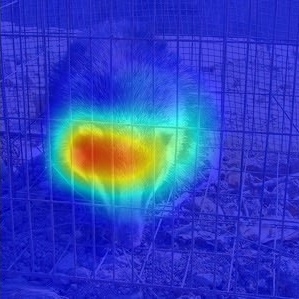}&
\includegraphics[width=0.55in]{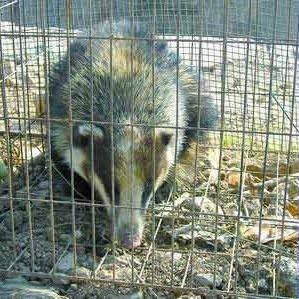}&
\includegraphics[width=0.55in]{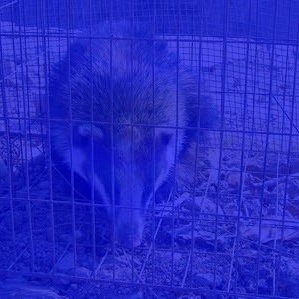}&
\includegraphics[width=0.55in]{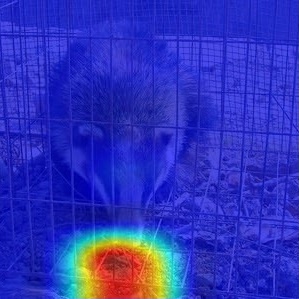} 
\\[-3pt]
 \multicolumn{2}{c}{\footnotesize $t_0$: badger } & & \multicolumn{2}{c}{\footnotesize $t$: computer }
\end{tabular}
\caption{\footnotesize{Visualizing CAMs of natural image and its adversarial  example (generated by C\&W attack) w.r.t. the true label `badger' and the target label `computer', respectively.  {The heat map color from blue to red represents the least and the most discriminative region localized by CAM, respectively. Here the values of CAMs  are normalized w.r.t. the largest value cross CAMs.}  
} }
\label{fig: CAM_examples_split}
\end{figure}

\textcolor{black}{
Let  $F(\mathbf x, c)$ denote the CAM for  image $\mathbf x$ w.r.t. the class label $c$.
The strength of a spatial element in  $F(\mathbf x, c)$  characterizes 
the importance of the activation at this spatial location for classifying $\mathbf x$ to the class $c$.
Thus,
one may wonder 
the relationship  between adversarial examples and discriminative regions localized by CAM. 
Given natural and adversarial examples,   CAMs of our interest include $F(\mathbf x_0, t_0)$, $F(\mathbf x^\prime, t_0)$, $F(\mathbf x_0, t)$ and $F(\mathbf x^\prime, t)$ with respect to both
  the correct and  the target labels; see \textcolor{black}{Figure\,\ref{fig: CAM_examples_split}} for an example and Figure\,\ref{fig: CAM_example} for more results.  Figure \ref{fig: CAM_examples_split} suggests that
the effect of adversarial
perturbations  can be visually explained through the class-specific discriminative image regions localized by CAM.
Compared $F(\mathbf x_0, t_0)$ with $F(\mathbf x^\prime, t_0)$,  the most discriminative region w.r.t. $(\mathbf x_0, t_0)$ is \textit{suppressed}  as   $\boldsymbol{\delta}$ is added to $\mathbf x_0$. By contrast, the difference between $F(\mathbf x_0, t)$ and $F(\mathbf x^\prime, t)$ implies that
the   discriminative region of  $\mathbf x_0$ under $t$ is \textit{enhanced}  after injecting   $\boldsymbol{\delta}$. 
}

We next employ   the so-called \textit{interpretability score} (IS) \citep{xu2018structured} to quantify how the adversarial perturbation   $\boldsymbol{\delta}$ is associated with the most discriminative image region found by the CAM  $F(\mathbf x, c)$.
More formally, let $B(\mathbf x, c)$ denote the Boolean map that highlights the most discriminative region,
{\small \begin{align}
     [B(\mathbf x,c)]_i = \left \{
     \begin{array}{ll}
       1   &  [F(\mathbf x,c)]_i \geq \nu \\
        0  &  \text{otherwise},
     \end{array}
     \right. \label{eq: B}
\end{align}}%
where $\nu > 0$ is a given threshold, and $[F(\mathbf x,c)]_i$ is the $i$th element of $F(\mathbf x,c)$. The IS of   adversarial perturbations   w.r.t. $(\mathbf x, c)$ is defined by 
{\small \begin{align}\label{eq: IS}
    \mathrm{IS}(\boldsymbol{\delta}) = {\| B(\mathbf x, c) \circ  \boldsymbol{\delta} \|_2}/{\| \boldsymbol{\delta} \|_2}.
\end{align}}%
where $\circ$ is the element-wise product. 
\textcolor{black}{Figure\,\ref{fig: v_check} shows the sensitivity of IS against the hyperparameter $\nu$. Not surprisingly, the threshold  $\nu$
cannot be too large or too small to  highlight  the proper discriminative image regions. We set  it as $70\%$-quantile  of weights in a CAM.}


\begin{figure}[b]
\centering 
\begin{adjustbox}{width=0.49\textwidth }
\begin{tabular}{cc}
\hspace{-3mm} \includegraphics[width=1.4in]{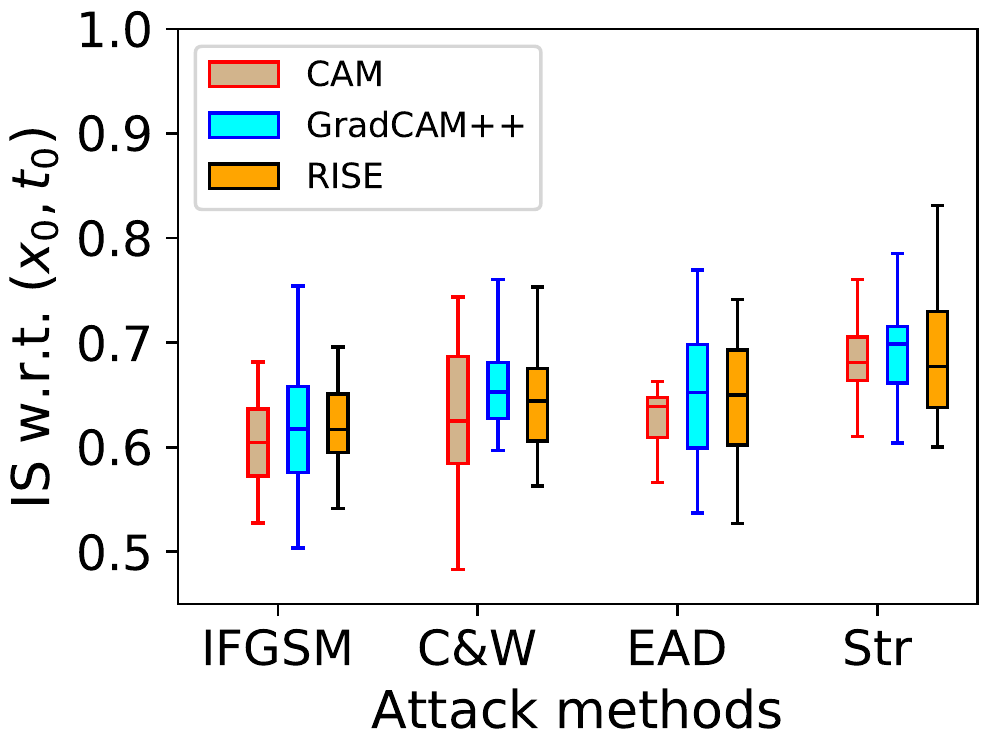}&
\hspace{-2mm}\includegraphics[width=1.4in]{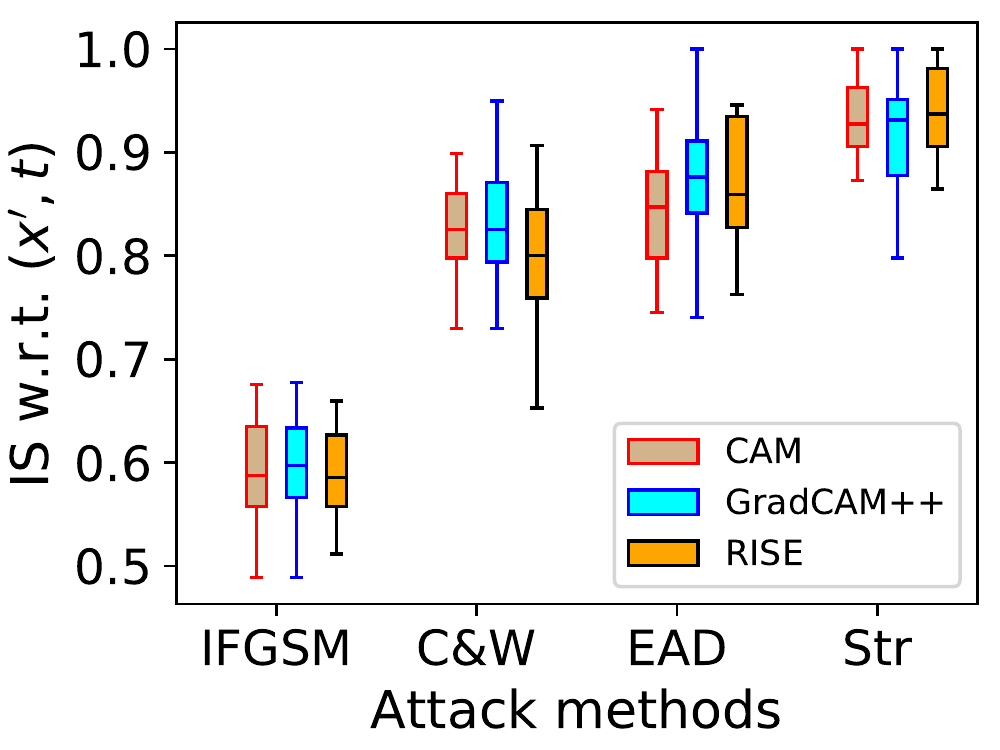}
\end{tabular}
\end{adjustbox}
\caption{IS  
under  $4$ attack types \& $3$ visual explanation methods on Resnet. Left: IS defined on $F(\mathbf x_0, t_0)$. Right: IS defined on $F(\mathbf x^\prime, t)$. Each box plot represents IS values of    $5000$ natural/adversarial examples from ImageNet.} 
\label{ISfigure}
\end{figure}

In \eqref{eq: IS},  
$\mathrm{IS}(\boldsymbol{\delta}) \to 1$ if the discriminative region   perfectly predicts  the locations of adversarial perturbations. By contrast, if $\mathrm{IS}(\boldsymbol{\delta}) \to 0$, then   adversarial perturbations cannot be interpreted by CAM. 
\textcolor{black}{In Figure\,\ref{ISfigure}, we    examine IS  for $4$   attack types via CAM, GradCAM++ and RISE w.r.t.      $(\mathbf x_0, t_0)$ and $(\mathbf x^\prime, t)$.   
We see that  IS  is  not  quite sensitive  to  the choice of interpretability methods, since it is built on   Boolean localizable maps, which enjoy  a large  overlapping among different visual explanation tools.
We also see that {Str} and IFGSM    yield  the best and the worst IS, respectively.
These results are consistent with Figure\,\ref{figure:correlation}: Str-attack tends to perturb  local semantic image regions, while IFGSM perturbs every pixel due to the use of sign-based perturbation direction.}

\begin{figure}[t]
   \centering
\begin{tabular}{p{0.1in}p{0.9in}p{0.9in}p{0.9in}}

 &  \hspace*{-0.03in} \footnotesize{adv. examples} & 
 \hspace*{-0.2in} \begin{tabular}[c]{@{}c@{}}\footnotesize{$F(\bm x_0 , t_0)$} \&  \footnotesize{PSRs}  \end{tabular} 
 &   \hspace*{-0.2in} 
  \begin{tabular}[c]{@{}c@{}}\footnotesize{ $F(\bm x^\prime , t)$} \& \footnotesize{PSRs}  \end{tabular}
 \\[-2pt]
\hspace*{-0.16in}  \rotatebox{90}{\parbox{1.in}{\centering \footnotesize suppression-\\ dominated effect}}  &

\hspace*{-0.15in}\includegraphics[width=1.05in]{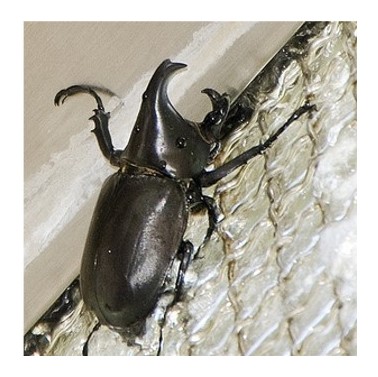}&
\hspace*{-0.2in}\includegraphics[width=1.05in]{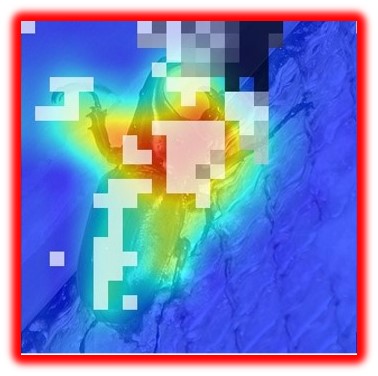}&
\hspace*{-0.25in}\includegraphics[width=1.05in]{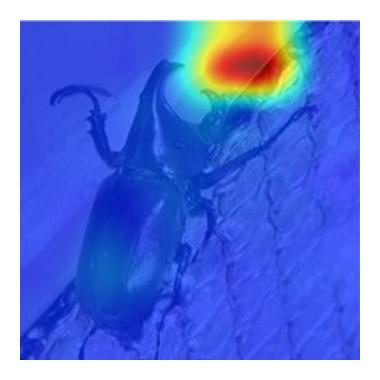} \\[-6pt]
& \multicolumn{3}{l}{\footnotesize{ $t_0$: rhinoceros beetle --  $t$: ambulance} }
\\[4pt]
\hspace*{-0.16in} \rotatebox{90}{\parbox{1.in}{\centering \footnotesize promotion-\\ dominated effect}} &
\hspace*{-0.15in}\includegraphics[width=1.05in]{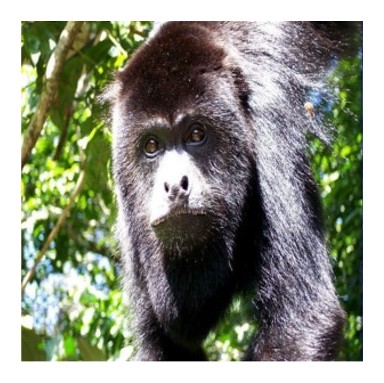}&
\hspace*{-0.2in}\includegraphics[width=1.05in]{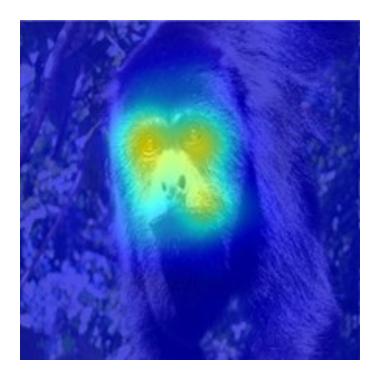}&
\hspace*{-0.25in}\includegraphics[width=1.05in]{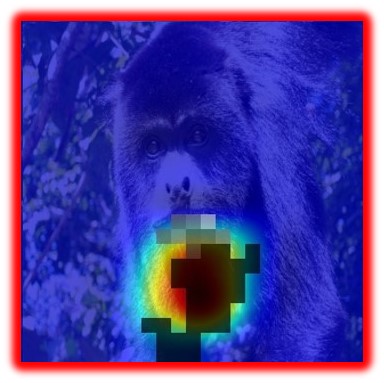}
\\[-8pt]
& \multicolumn{3}{l}{\footnotesize{$t_0$: howler monkey --  $t$: paper towel} }
\\[4pt]
\hspace*{-0.16in} \rotatebox{90}{\parbox{1.in}{\centering \footnotesize balance-\\dominated effect}} &
\hspace*{-0.15in}\includegraphics[width=1.05in]{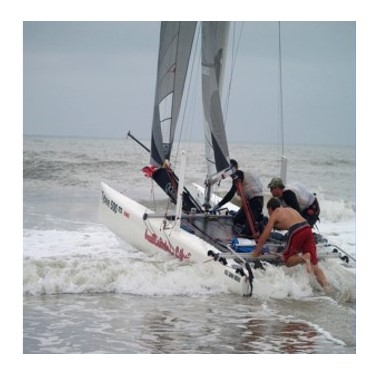}&
\hspace*{-0.2in}\includegraphics[width=1.05in]{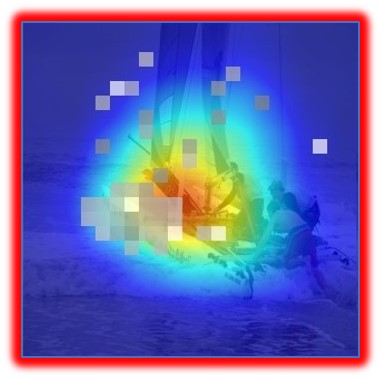}&
\hspace*{-0.25in}\includegraphics[width=1.05in]{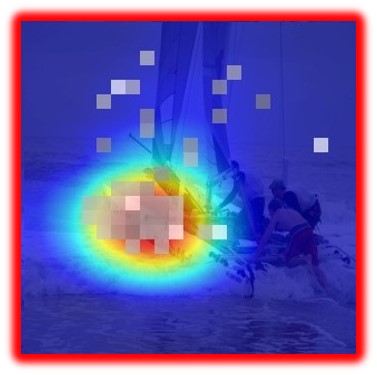}\\[-6pt]
& \multicolumn{3}{l}{ \footnotesize{ $t_0$: catamaran -- $t$: container ship}}
\\
\end{tabular}
\caption{\footnotesize{Interpreting adversarial perturbations via CAM and PSR.  
For PSR, only the top $70\%$ most significant   perturbed grids {ranked by   $\{s_i\}$ \eqref{eq: s_d}} are shown. The white and black colors represent
the suppression-dominated regions   ($r_i < -1$) and the promotion-dominated regions  ($r_i > 1$), respectively. The gray color  corresponds to balance-dominated perturbations  ($r_i \in [-1,1]$). And the red box represents the dominated adversarial effect. 
}}
    \label{fig: CAM_example_grid}
\end{figure}

\begin{figure}[t]
\centerline{
\begin{tabular}{cccc}
adv. & \hspace*{-0.15in}\footnotesize{$F(\bm x^\prime , t)$ \& PSR}  & adv. &  \hspace*{-0.15in}\footnotesize{$F(\bm x^\prime , t)$ \& PSR}               \\
\includegraphics[width=.11\textwidth]{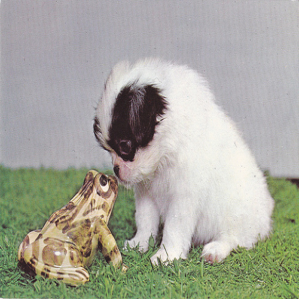}  
&\hspace*{-0.15in} \includegraphics[width=.11\textwidth]{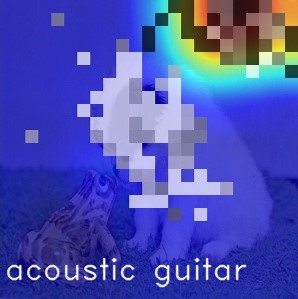}
& \includegraphics[width=.11\textwidth]{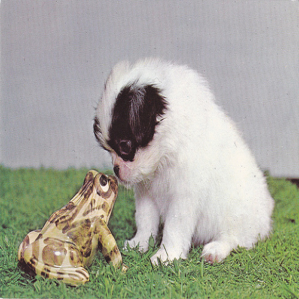}
& \hspace*{-0.1in}\includegraphics[width=.11\textwidth]{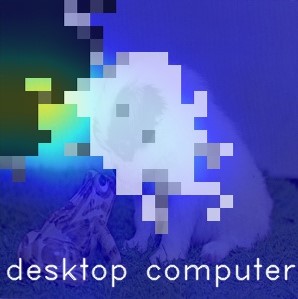}
\end{tabular}}
\caption {\footnotesize{Interpreting adversarial examples of the original image   `Japanese spaniel' in Figure\,\ref{fig: intro} w.r.t. different target labels `acoustic guitar' and `desktop computer'  using CAM and PSR. 
}}
  \label{fig: dog_2}
\end{figure}

\textcolor{black}{
We recall that PSR in \eqref{eq: ri}
categorizes
$\{\boldsymbol{\delta}_{\mathcal G_i} \}$ into three types: 
 suppression-dominated perturbations,  promotion-dominated perturbations, and balanced perturbations. 
  In Figure\,\ref{fig: CAM_example_grid}, we see that the locations and the promotion-suppression roles of adversarial perturbations are well matched to the  discriminative regions of $F(\mathbf x_0, t_0)$ and $F(\mathbf x^\prime, t)$. 
 \textcolor{black}{In particular, 
 if there exists a large overlapping between  $F(\mathbf x_0, t_0)$ and $F(\mathbf x^\prime, t)$, then
 the balanced perturbations are desired 
 since perturbing a single pixel can play a dual role in suppression and promotion.  
 Toward deeper insights, we investigate how the adversary makes an impact on attacking a single image with multiple target labels (Figure\,\ref{fig: dog_2}) as well as  attacking multiple images with the same source and target label (Figure\,\ref{fig: onevsone}).
 We see that 
 the promotion-dominated perturbation  is adaptive  to the change of the target label in Figure\,\ref{fig: dog_2}.
Moreover, the same source-target label pair
enforces a similar effect of  adversarial perturbations on attacking different images in Figure\,\ref{fig: onevsone}.
Additional results can be found in Appendix\,\ref{app: adv_example_cam}.
 }
}

\textit{Insights on how CAM   constrains  effectiveness of adversarial attacks.}
We have previously shown that
CAM can be used to localize  class-specific discriminative image regions. 
We now consider two types of CAM-based operations to refine an adversarial perturbation pattern:
  (a) removing less sensitive perturbations  quantified by  $s_i$ in \eqref{eq: s_d}, and (b) enforcing perturbations  in the   most  discriminative  regions  only w.r.t. the true label, namely, $B(\mathbf x_0, t_0)$ in \eqref{eq: B}. 
  \textcolor{black}{We represent  the  refinement operations  (a) and (b) through the constraint sets of pixels
$\mathcal S_1 = \{ \forall i \, | \, s_i > \beta  \}$ for a  positive threshold $\beta$ and $\mathcal S_2 = \{ \forall i \, | \, [B(\mathbf x_0, t_0)]_i > 0 \}$, where $\beta$ is set to   filter perturbations of less than $1 - \nu = 30\%$ cumulative  sensitivity scores. The refined adversarial examples are then generated by performing the existing attack methods with an additional projection on the sparse constraints given by $\mathcal S_1$ and $\mathcal S_2$. We refer readers to Appendix\,\ref{app: attack_refine} for more details.}

\begin{figure}[hb]
   \centering
 \begin{tabular}{p{0.8in}p{0.8in}p{0.9in}}
 \hspace*{0.1in} \begin{tabular}[c]{@{}c@{}}\scriptsize{original} \vspace*{-0.11in}  \end{tabular} 
 &  \hspace*{-0.35in} \parbox{1.2in}{\centering \scriptsize{CAM + PSRs  w.r.t.} \\ \scriptsize adv. image \& target label \vspace*{0.02in}}  
 & \hspace*{-0.2in} \parbox{1.0in}{ \centering \scriptsize perturbations \\ \scriptsize (Str-attack)} 
\\
\hspace*{-0.2in} \includegraphics[width=0.9in]{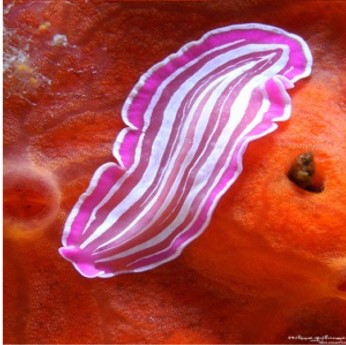} & \hspace*{-0.2in}
 \includegraphics[width=0.9in]{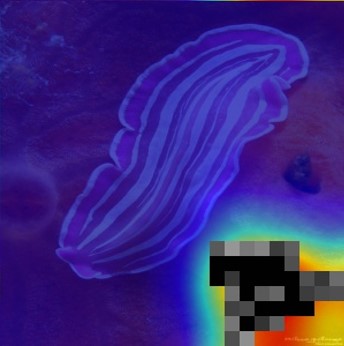} & \hspace*{-0.2in}
\includegraphics[width=1.2in,height = 0.91in]{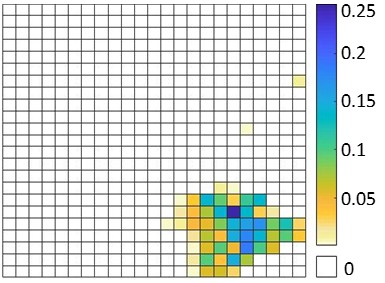}
 \\[0.1cm]
 \hspace*{-0.2in} \parbox{0.9in}{\centering \scriptsize{CAM  w.r.t.} \\ \scriptsize ori. image \& true label}
 &  \hspace*{-0.25in} \parbox{1.1in}{\centering \scriptsize  CAM  + PSRs w.r.t.\\ \scriptsize  refined attack \& true label}  &  
 \hspace*{-0.15in} \parbox{1.0in}{ \centering \scriptsize perturbations \\ \scriptsize (refined Str-attack)} 
 \\
\hspace*{-0.2in} \includegraphics[width=0.9in]{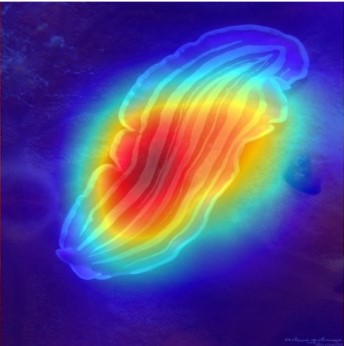} & \hspace*{-0.2in}
\includegraphics[width=0.9in]{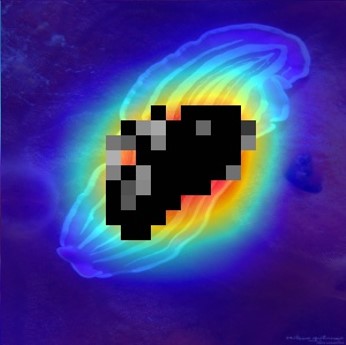} & \hspace*{-0.2in}
\includegraphics[width=1.2in]{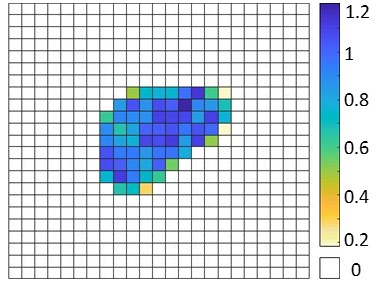}
\end{tabular}
\caption{\footnotesize{
\textcolor{black}{ 
The `flatworm'-to-`knot' adversarial example (generated by str-Attack) with  and without refinement under  $\mathcal S_2$. The first row presents the original image, PSRs overlaid on CAM of the adversarial example w.r.t. the target label `knot', and the $\ell_2$-norm distortion of adversarial perturbations. The second row presents $\mathcal S_2$ given by CAM of the original image w.r.t. the true label `flatworm', and the refined attack under $\mathcal S_2$. Note that this refinement leads to much larger $\ell_2$ distortion  (max. value  $1.2$) against the unrefined attack (max. value  $0.25$); see the third column.}
}}
    \label{fig: refine2}
\end{figure}

We find that  it is possible to obtain a more effective attack by perturbing  much less    pixels of high sensitivity scores under $\mathcal S_1$, but without increasing  $\ell_p$  perturbation strength (Table\,\ref{table:refine}).
For attacks with refinement  under $\mathcal S_2$,  
Figure\,\ref{fig: refine2} shows that perturbing pixels  under only a suppression-dominated  adversarial pattern  $\mathcal S_2 $  w.r.t. the true label
is \textit{not} effective: If we restrict perturbations under  $\mathcal S_2$,  then  the refined attack leads to a much larger $\ell_2$ distortion. That is because  the  perturbation $\boldsymbol{\delta}$ 
originally plays a role in   promoting the confidence of the target label, which corresponds to a  class-specific discriminative region different  from $\mathcal S_2$. 

\section{\textcolor{black}{Seeing Effects of Adversarial  Perturbations from Network Dissection}}
\label{sec: network}

In this section, we examine the promotion-suppression effect of adversarial perturbations on the internal response of CNNs by leveraging network dissection  \citep{bau2017network}. 
We show that there exists a  connection between the sensitivity of units (a unit refer to a channel-wise feature map) and their concept-level interpretability. 

We  begin by reviewing the main   idea of network dissection; see more details in \citep{bau2017network}. 
Interpretability measured by   network dissection    refers to  the alignment between individual hidden units and a set of semantic concepts provided by the   broadly and densely labeled dataset \textit{Broden}. 
Different from other datasets, examples in Broden contain 
 pixel-level concept annotation, ranging from low-level concepts such as \textit{color} and \textit{texture} to higher-level concepts such as \textit{material}, \textit{part}, \textit{object} and \textit{scene}. Network dissection  builds a correspondence between  a hidden unit's activation    and its interpretability on semantic concepts. More formally, the interpretability of unit (IoU) $k$ w.r.t. the concept $c$ is defined by \citep{bau2017network} 
 {\small \begin{align}\label{eq: IOU_standard}
     \mathrm{IoU}(k,c) = \frac{\sum_{\mathbf x \in \mathcal{D}} |  \mathbf M_k(\mathbf x) \cap \mathbf L_c(\mathbf x) | }{ \sum_{\mathbf x \in \mathcal{D}} |  \mathbf M_k(\mathbf x) \cup \mathbf L_c(\mathbf x) |  },
 \end{align}}%
 where $\mathcal D$ denotes Broden, and $|\cdot|$ is the cardinality of a set. In \eqref{eq: IOU_standard}, $\mathbf M_k(\mathbf x)$ is a binary segmentation of the activation map of unit $k$, which gives the representative region of $\mathbf x$ at $k$. Here the activation is  scaled up  to  the input resolution using  bilinear interpolation, denoted by $\mathbf S_k (\mathbf x)$, and  then truncated using the top $5\%$ quantile
 (dataset-level) threshold $T_k$. That is, $\mathbf M_k(\mathbf x) = \mathbf S_k (\mathbf x) \geq T_k$, namely, 
 the $(i,j)$th element of $\mathbf M_k(\mathbf x)$ is $1$ if the $(i,j)$th element of $\mathbf S_k (\mathbf x)$ is greater than or equal to $T_k$, and $0$ otherwise.
 In \eqref{eq: IOU_standard}, $\mathbf L_c(\mathbf x)$ is the input-resolution annotation mask, provided by Broden, for the concept $c$ w.r.t. $\mathbf x$. Since 
 one unit might  detect multiple concepts, 
 the   interpretability of a unit is summarized as 
$\mathrm{IoU}(k) = (1/|\mathcal C|)\sum_{c} \mathrm{IoU}(k,c)$, where $|\mathcal C|$ denotes the total number of concept labels.
 
We next investigate the effect of adversarial perturbations on the internal response of CNNs by leveraging network dissection. We produce  adversarial examples $\mathcal D^\prime$ from  Broden  using the PGD untargeted attack method \citep{madry2017towards}.
Given    adversarial examples $\{ \mathbf x^\prime \in \mathcal D^\prime \}$, we characterize the sensitivity of unit $k$  (to 
 adversarial perturbations) via 
the  change of activation segmentation 
{\small \begin{align}\label{eq: v_sens}
v(k) \Def  \mathbb E_{(\mathbf x, \mathbf x^\prime )} \left [
\left   \| \mathbf M_k(\mathbf x)  -   \mathbf M_k(\mathbf x^\prime ) \right \|_2 
\right ],
\end{align}}%
where $(\mathbf x, \mathbf x^\prime)$ is a pair of natural and adversarial examples, and the expectation is taken over a certain distribution of our interest, e.g., the entire dataset or data of  fixed source-target labels.
In  \eqref{eq: v_sens}, we adopt the  activation segmentation  $\mathbf M_k$ rather than the activation map $\mathbf S_k$ since the former highlights the representative region of an activation map without   inducing the layer-wise magnitude bias.

Given the per-unit sensitivity measure $v(k)$ and interpretability measure $\mathrm{IoU}(k)$, 
we may ask \textit{whether or not the  sensitive units (to adversarial perturbations) exhibit strong interpretability.} To answer this question, we conduct the statistical significance test 
by contrasting the IoU of the top $N$ ranked sensitive units with the IoU distribution of   randomly selected $N$ units. Formally, the $p$-value is the probability of observing $\sum_{k} \mathrm{IoU}(k)$ when $k$ is from  top $N$  sensitive units ranked by $v(k)$  in the background IoU distribution      when $N$ units are randomly picked.
The smaller the $p$-value is, the more significant the connection between sensitivity and interpretability is.

\begin{figure}[t]
\centerline{
\hspace*{-0.05in} \begin{tabular}{cc}
      \includegraphics[width=.241\textwidth]{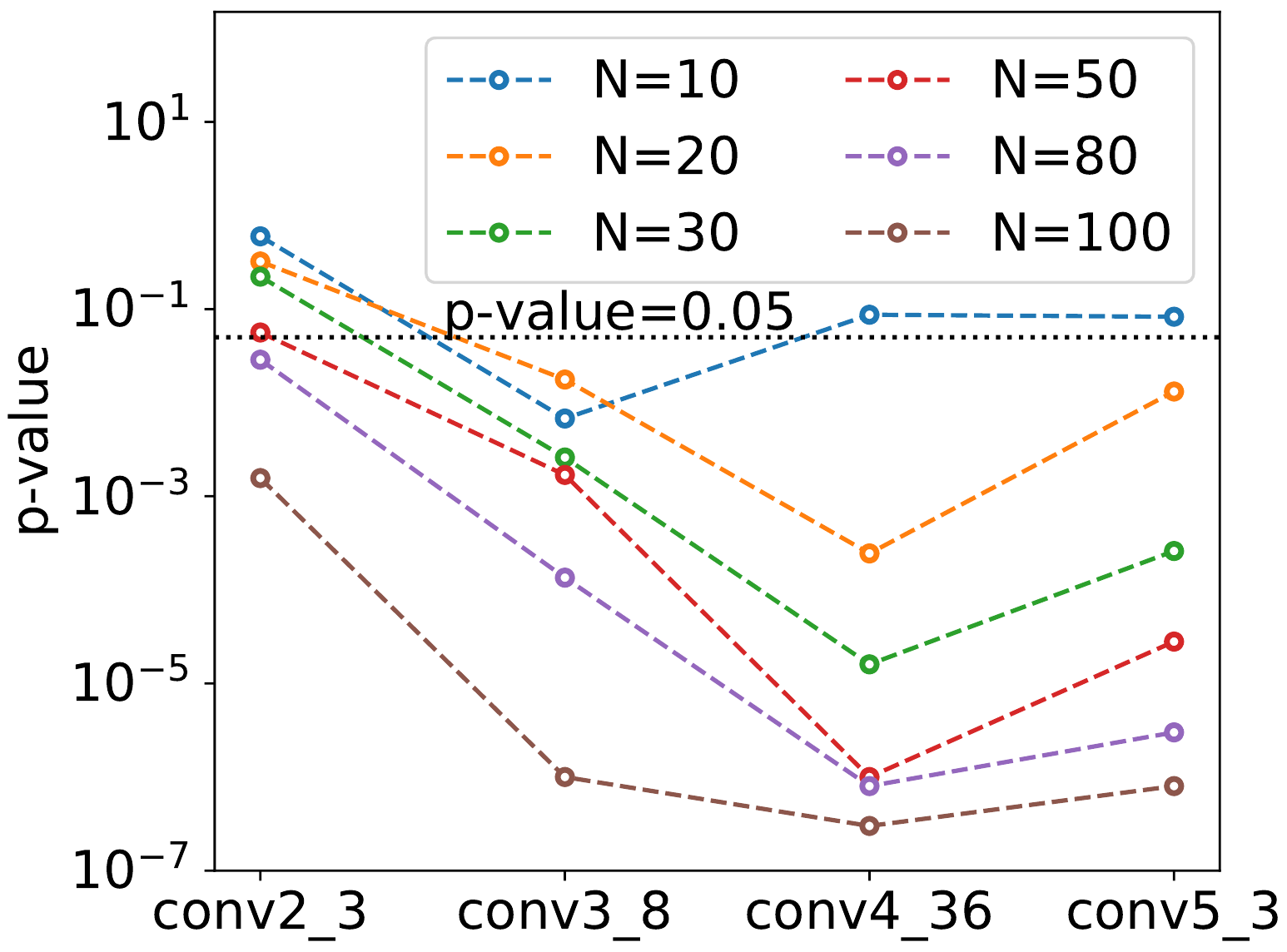}   & \hspace*{-0.18in}
       \includegraphics[width=.2391\textwidth]{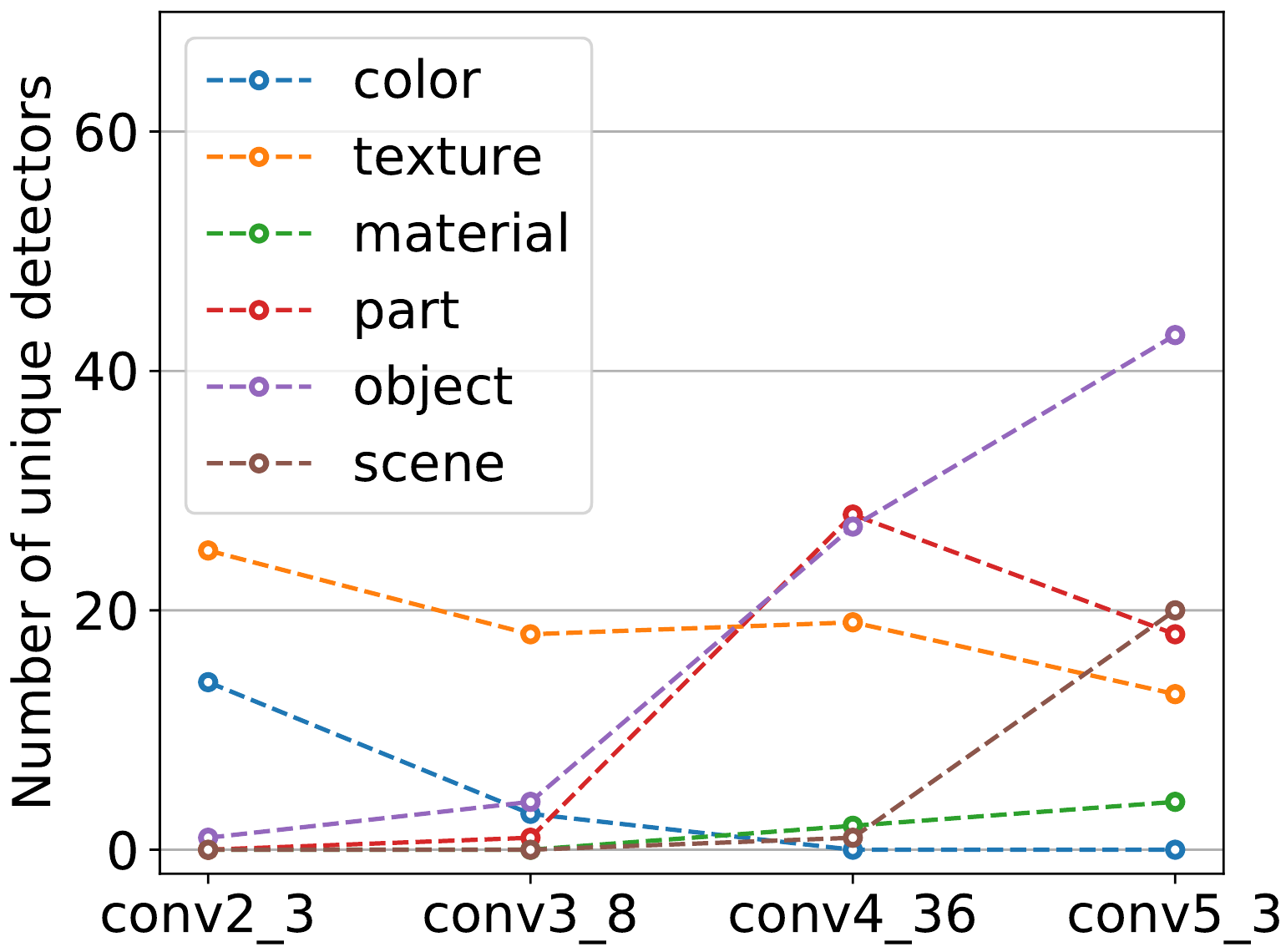}       \vspace*{-0.05in} \\
      \footnotesize{(a)} & \hspace*{-0.18in} \footnotesize{(b)}
            \vspace*{-0.1in}
\end{tabular}}
\caption{\footnotesize{Sensitivity and interpretability. (a) $p$-value of interpretability of top $N$ sensitive units to adversarial attacks in Resnet\_$152$, where the presented  layers include  conv2\_3 (256 units), conv3\_8 (512 units), conv4\_36 (1024 units) and conv5\_3 (2048 units).   (b) Number of concept detectors among top $N = 100$  sensitive units per layer for each concept category. 
}} 
  \label{fig: pvalue_concept} 
 \vspace*{-4mm}
\end{figure}

\textcolor{black}{
We present the significance test of the interpretability of top $N \in \{ 10, 20, 30, 50, 80, 100\}$   sensitive units against the layer index of Resnet\_$152$ (Figure\,\ref{fig: pvalue_concept}-a). We also show the number of concept detectors\footnote{A concept detector refers to a unit with the top ranked concept satisfying $\max_c \mathrm{IoU}(k,c) > 0.04$ \citep{bau2017network}. 
}  among top $N =100$ sensitive units versus  layers for every concept category (Figure\,\ref{fig: pvalue_concept}-b). 
Here we denote by conv$i\_j$   the last convolutional layer of $j$th building block at the $i$th layer in Resnet\_$152$ \citep{he2016deep}.
It is seen from Figure\,\ref{fig: pvalue_concept}-a  that there exists a strong connection between the sensitivity of units and their   interpretability since $p < 0.05$ in most of cases. 
By fixing the layer number,  such a connection becomes more significant as $N$ increases:
Most of the top $100$ sensitive   units are interpretable, although the top $10$ sensitive units might not be the same top $10$ interpretable units.
By fixing $N$, we observe that deep layers (conv4\_36 and conv5\_3) exhibit stronger    connection between sensitivity and interpretability compared to shallow layers (conv2\_3 and conv3\_8). 
That is because
 the change of activation induced by adversarial attacks at shallow layers could be subtle  and are less detectable in terms of interpretability. Indeed, Figure\,\ref{fig: pvalue_concept}-b shows that
  more high-level concept detectors (e.g., object and part)
  emerge in conv4\_36 and conv5\_3 while  low-level concepts (e.g., color and texture)
dominate at lower layers.
}

\begin{figure*}[htb]
  \centering
  \begin{adjustbox}{max width=1\textwidth }
  \begin{tabular}{@{\hskip 0.00in}c  @{\hskip 0.00in} @{\hskip 0.02in} c @{\hskip 0.02in} | @{\hskip 0.02in} c @{\hskip 0.02in} |@{\hskip 0.02in} c @{\hskip 0.02in} | @{\hskip 0.02in} c@{\hskip 0.02in}  }
& 
\colorbox{light-gray}{\footnotesize \textbf{conv2\_3}} 
&  
\colorbox{light-gray}{\footnotesize \textbf{conv3\_8}} 
&
\colorbox{light-gray}{ \footnotesize \textbf{conv4\_36}}
&  
\colorbox{light-gray}{\footnotesize \textbf{conv5\_3}}
\\
 \begin{tabular}{@{}c@{}}  
\vspace*{0.1in}\\
\rotatebox{90}{\parbox{5em}{\centering \footnotesize \textbf{Ori}: table lamp ($t_0$)}}
 \vspace*{-0.05in}
 \\
\rotatebox{90}{\parbox{5em}{\centering \footnotesize \textbf{Adv}: studio couch ($t$)}}
\end{tabular} 
&
\begin{tabular}{@{\hskip 0.02in}c@{\hskip 0.02in}}
     \begin{tabular}{@{\hskip 0.00in}c@{\hskip 0.00in}}
     \parbox{10em}{\centering \footnotesize  unit193, orange-color}  
    \end{tabular} 
    \\
 \parbox[c]{10em}{\includegraphics[width=10em]{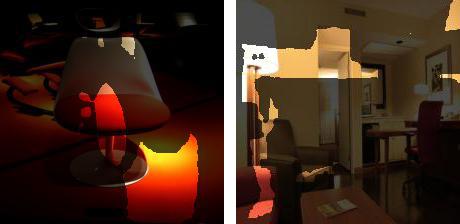}} 
 \\
 \parbox[c]{10em}{\includegraphics[width=10em]{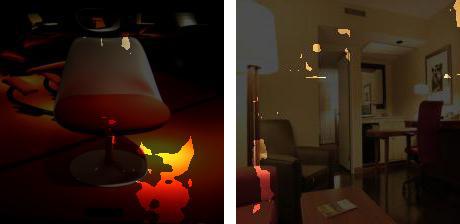}} 
\end{tabular}
&
 \begin{tabular}{@{\hskip 0.02in}c@{\hskip 0.02in}} 
      \begin{tabular}{@{\hskip 0.00in}c@{\hskip 0.00in}}
     \parbox{10em}{\centering \footnotesize unit358, flecked-texture}  
    \end{tabular} 
    \\
 \parbox[c]{10em}{\includegraphics[width=10em]{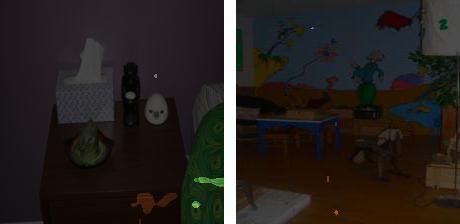}} 
 \\
 \parbox[c]{10em}{\includegraphics[width=10em]{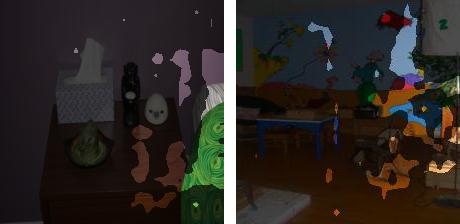}} 
\end{tabular}
&
 \begin{tabular}{@{\hskip 0.02in}c@{\hskip 0.02in}} 
      \begin{tabular}{@{\hskip 0.00in}c@{\hskip 0.00in}}
     \parbox{10em}{\centering \footnotesize  unit457, shade-part}   
    \end{tabular} 
    \\
 \parbox[c]{10em}{\includegraphics[width=10em]{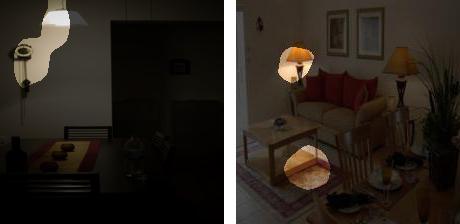}} 
 \\
 \parbox[c]{10em}{\includegraphics[width=10em]{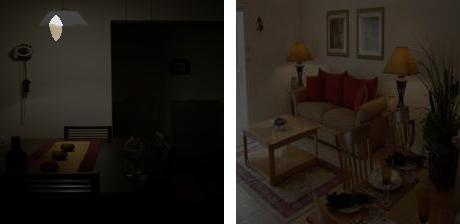}} 
\end{tabular}
&
 \begin{tabular}{@{\hskip 0.02in}c@{\hskip 0.02in}c@{\hskip 0.02in} } 
      \begin{tabular}{@{\hskip 0.00in}c@{\hskip 0.00in}}
     \parbox{10em}{\centering \footnotesize  unit1716, lamp-object}   
    \end{tabular} 
     &  
      \begin{tabular}{@{\hskip 0.00in}c@{\hskip 0.00in}}
       \parbox{10em}{\centering \footnotesize  unit123, sofa-object}   
    \end{tabular} 
    \\
 \parbox[c]{10em}{\includegraphics[width=10em]{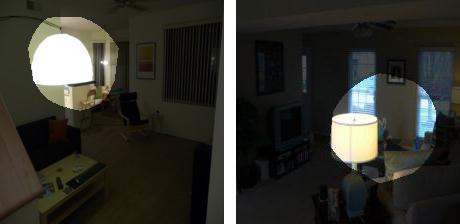}} &    \parbox[c]{10em}{\includegraphics[width=10em]{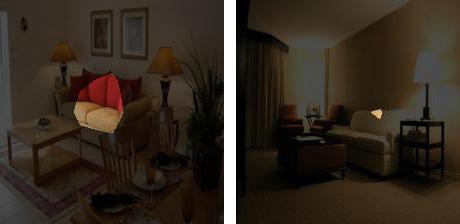}} 
 \\
 \parbox[c]{10em}{\includegraphics[width=10em]{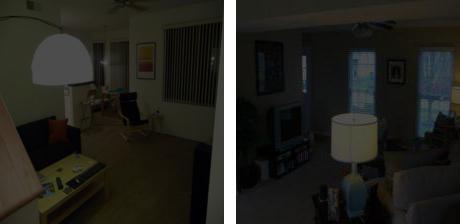}} &    \parbox[c]{10em}{\includegraphics[width=10em]{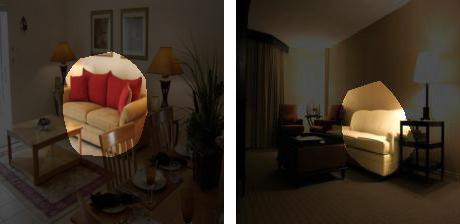}} 
\end{tabular}\\
\vspace*{-0.165in}\\
 \begin{tabular}{@{}c@{}}  
\vspace*{0.00in}\\
\rotatebox{90}{\parbox{3.8em}{\centering \footnotesize \textbf{Ori}:  airliner ($t_0$)}}
 \vspace*{0.05in}
 \\
\rotatebox{90}{\parbox{4em}{\centering \footnotesize  \textbf{Adv}:  seashore ($t$)}}
\end{tabular} 
&
\begin{tabular}{@{\hskip 0.02in}c@{\hskip 0.02in}} 
     \begin{tabular}{@{\hskip 0.00in}c@{\hskip 0.00in}}
      \parbox{10em}{\centering \footnotesize  unit84, blue-color}   
    \end{tabular} 
    \\
 \parbox[c]{10em}{\includegraphics[width=10em]{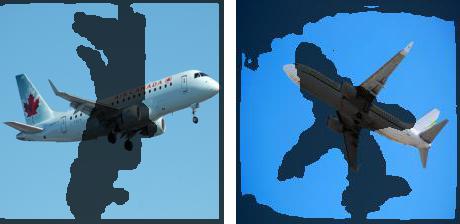}} 
 \\
 \parbox[c]{10em}{\includegraphics[width=10em]{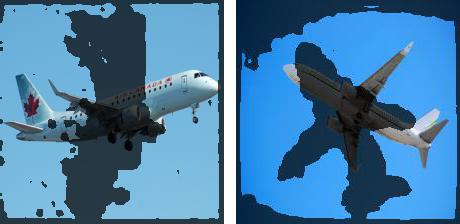}}
\end{tabular}
&
 \begin{tabular}{@{\hskip 0.02in}c@{\hskip 0.02in}} 
      \begin{tabular}{@{\hskip 0.00in}c@{\hskip 0.00in}}
     \parbox{10em}{\centering \footnotesize  unit445, banded-texture}   
    \end{tabular} 
    \\
 \parbox[c]{10em}{\includegraphics[width=10em]{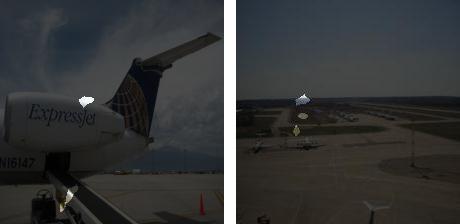}}
 \\
 \parbox[c]{10em}{\includegraphics[width=10em]{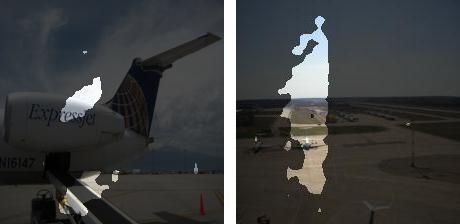}}
\end{tabular}
&
 \begin{tabular}{@{\hskip 0.02in}c@{\hskip 0.02in}} 
      \begin{tabular}{@{\hskip 0.00in}c@{\hskip 0.00in}}
     \parbox{10em}{\centering \footnotesize  unit2, stern-part}   
    \end{tabular} 
    \\
 \parbox[c]{10em}{\includegraphics[width=10em]{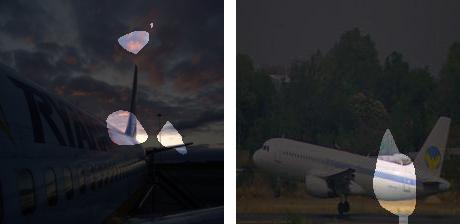}} 
 \\
 \parbox[c]{10em}{\includegraphics[width=10em]{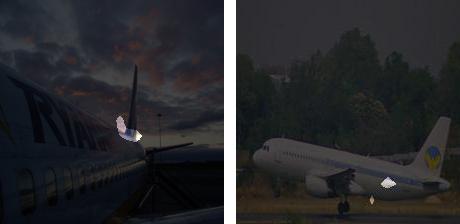}}
\end{tabular}
&
 \begin{tabular}{@{\hskip 0.02in}c@{\hskip 0.02in}c@{\hskip 0.02in}} 
      \begin{tabular}{@{\hskip 0.00in}c@{\hskip 0.00in}}
     \parbox{10em}{\centering \footnotesize  unit781, airplane-object}   
    \end{tabular} 
     &  
      \begin{tabular}{@{\hskip 0.00in}c@{\hskip 0.00in}}
     \parbox{10em}{\centering \footnotesize  unit782, beach-scene}   
    \end{tabular} 
    \\
 \parbox[c]{10em}{\includegraphics[width=10em]{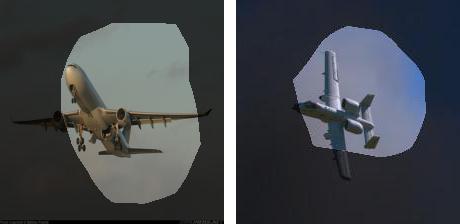}} &    \parbox[c]{10em}{\includegraphics[width=10em]{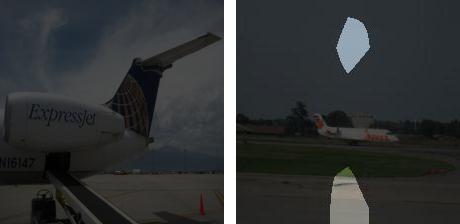}} 
 \\
 \parbox[c]{10em}{\includegraphics[width=10em]{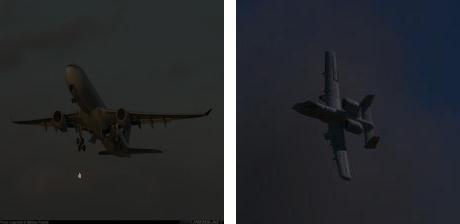}} &    \parbox[c]{10em}{\includegraphics[width=10em]{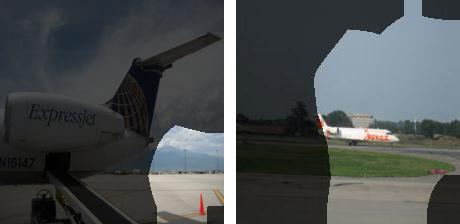}}
\end{tabular}
\vspace*{-0.01in}
\end{tabular}
  \end{adjustbox}
    \caption{\footnotesize{Visualizing impact of original (Ori) \& adversarial (Adv) examples on the response of concept detectors identified by network dissection at $4$ representative  layers in Resnet.   
   (top) attack `table lamp'-to-`studio couch, day bed', (bottom) attack `airliner'-to-`seashore, seacoast'.   In both top and bottom sub-figures, the first row presents   unit indices together with   detected top ranked concept labels and categories (in the format   `concept label'-`concept category'). The last two rows present the response of concept detectors visualized by the segmented input image, where the segmentation is given by $\mathbf M_k(\mathbf x)$ corresponding to the  top ranked concept at each unit. 
   }}
  \label{fig: dissec}
 \vspace*{-4mm}
\end{figure*}

To peer into the impact of adversarial perturbations on individual   images,
 we examine how the representation of concept detectors change while facing adversarial examples  by attacking images from the same true class $t_0$ to  the same target class $t$. Here the representation of a  concept detector is visualized by the segmented input image determined by $\mathbf M_k(\mathbf x)$.  
In Figure\,\ref{fig: dissec}, we show two examples of  attacks: `table lamp'-to-`studio couch, day bed' and `airliner'-to-`seashore, seacoast'. We first note that most of low-level concepts (e.g.,  color and texture) are detected at shallow layers, consistent with Figure\,\ref{fig: pvalue_concept}-b.  
In the attack `table lamp'-to-`studio couch, day bed', the color `orange'  detected at    conv2\_3  is less expressed for the adversarial image against the natural image. This aligns with human perception since `orange' is  related to `light' and thus `table lamp'.
By contrast, in the attack `airliner'-to-`seashore, seacoast',    the color `blue' is well detected at both natural and adversarial images, since `blue' is associated with both `sky' for `airliner' and `sea' for `seashore'.
We also note  that high-level concepts (e.g., part and object) dominate at deeper layers. At  conv5\_3,  the expression of object concepts  (e.g., lamp and airplane)  relevant to the \textit{true} label is \textit{suppressed}. Meanwhile, the expression of object concepts (e.g., sofa and beach)  relevant to the \textit{target} label is \textit{promoted}. This precisely reflects the activation promotion-suppression effect induced by adversarial perturbations. \textcolor{black}{In Figure\,\ref{fig: CAM_Netdisecction}, we connect images in Figure\,\ref{fig: dissec} to PSR and CAM based visual explanation.}

\section{{Insights of Interpretability  for Improving Adversarial    Robustness}}


\textit{First, PSE  explains the  effectiveness of detecting adversarial examples with feature attribution \citep{yang2019ml}.} 
It was   shown in \citep{yang2019ml} that the input attribution scores of an adversarial example obtained by the leave-one-out method \citep{zeiler2014visualizing} yields a significantly larger variance than the case of natural image.  The presence of a significant change on  the probability of the top-$1$ class,  when a   pixel is removed from an adversarial example,
is explainable. By PSE, the possible reason is that the seemingly random perturbation could play a critical role in promoting the confidence of the \textit{target} label, e.g., Figure\,\ref{fig: dog_2}.  

\textit{Second, hiding adversarial examples from CAM may be not easy.}
It was shown in \citep{zhang2018interpretable} that adversarial examples can be crafted to fool CNNs, and at the same time keep their CAMs (w.r.t. the top-1 prediction class) intact. However, our results in Sec.\,\ref{sec: image} suggest that the   discrepancy between CAMs of natural and adversarial examples exists w.r.t. \textit{both}   the true   and   the target label.
Thus, we need to re-think whether or not it is  easy to hide adversarial examples from network interpretation defined under the two-class or even all-class CAM distortion.

\textit{Third,   network dissection  implies the method of neuron masking to improve robustness.}
Since the sensitive units to adversarial perturbations exhibit strong interpretability,
one could mitigate  the  effect of adversaries by   masking these   sensitive neurons with interpretation toward the target label.  Our preliminary results in Table\,\ref{table:correlation} of Appendix\,\ref{app: dissection} show that the suggested neuron masking improves robustness  at the cost of slight degradation on clean test accuracy.


\section{Conclusions}
\textcolor{black}{
In this work, we made a significant effort to understand the mechanism of adversarial attacks and provided its   explanation  at pixel, image and network levels.
We showed that  adversarial attacks play a significant role in activation promotion and suppression.
The promotion-suppression effect is  strongly associated with 
class-specific discriminative  image  regions. 
We also demonstrated that the  interpretable adversarial pattern  constrains the effectiveness of adversarial attacks. 
We further provided the first analysis of adversarial examples through network dissection, which builds the 
connection between the units' sensitivity to imperceptible perturbations and    their interpretability on semantic concepts. 
In the future, we would like to develop interpretability-driven defensive methods and consider the scenario of attack against interpretability, not just prediction. 
}

{\small
\clearpage \pagebreak
\bibliography{egbib}

\begin{thebibliography}{}

\bibitem[\protect\citeauthoryear{Athalye, Carlini, and
  Wagner}{2018}]{athalye2018obfuscated}
Athalye, A.; Carlini, N.; and Wagner, D.
\newblock 2018.
\newblock Obfuscated gradients give a false sense of security: Circumventing
  defenses to adversarial examples.
\newblock {\em arXiv preprint arXiv:1802.00420}.

\bibitem[\protect\citeauthoryear{Bau \bgroup et al\mbox.\egroup
  }{2017}]{bau2017network}
Bau, D.; Zhou, B.; Khosla, A.; Oliva, A.; and Torralba, A.
\newblock 2017.
\newblock Network dissection: Quantifying interpretability of deep visual
  representations.
\newblock {\em arXiv preprint arXiv:1704.05796}.

\bibitem[\protect\citeauthoryear{Bau \bgroup et al\mbox.\egroup
  }{2019}]{bau2018visualizing}
Bau, D.; Zhu, J.-Y.; Strobelt, H.; Zhou, B.; Tenenbaum, J.~B.; Freeman, W.~T.;
  and Torralba, A.
\newblock 2019.
\newblock Visualizing and understanding generative adversarial networks.
\newblock In {\em International Conference on Learning Representations}.

\bibitem[\protect\citeauthoryear{Boyd \bgroup et al\mbox.\egroup
  }{2011}]{boyd2011distributed}
Boyd, S.; Parikh, N.; Chu, E.; Peleato, B.; Eckstein, J.; et~al.
\newblock 2011.
\newblock Distributed optimization and statistical learning via the alternating
  direction method of multipliers.
\newblock {\em Foundations and Trends{\textregistered} in Machine Learning}
  3(1):1--122.

\bibitem[\protect\citeauthoryear{Brown \bgroup et al\mbox.\egroup
  }{2017}]{brown2017adversarial}
Brown, T.~B.; Man{\'e}, D.; Roy, A.; Abadi, M.; and Gilmer, J.
\newblock 2017.
\newblock Adversarial patch.
\newblock {\em arXiv preprint arXiv:1712.09665}.

\bibitem[\protect\citeauthoryear{Carlini and
  Wagner}{2017a}]{carlini2017adversarial}
Carlini, N., and Wagner, D.
\newblock 2017a.
\newblock Adversarial examples are not easily detected: Bypassing ten detection
  methods.
\newblock In {\em Proceedings of the 10th ACM Workshop on Artificial
  Intelligence and Security},  3--14.
\newblock ACM.

\bibitem[\protect\citeauthoryear{Carlini and
  Wagner}{2017b}]{carlini2017towards}
Carlini, N., and Wagner, D.
\newblock 2017b.
\newblock Towards evaluating the robustness of neural networks.
\newblock In {\em Security and Privacy (SP), 2017 IEEE Symposium on},  39--57.
\newblock IEEE.

\bibitem[\protect\citeauthoryear{Carlini \bgroup et al\mbox.\egroup
  }{2016}]{carlini2016hidden}
Carlini, N.; Mishra, P.; Vaidya, T.; Zhang, Y.; Sherr, M.; Shields, C.; Wagner,
  D.; and Zhou, W.
\newblock 2016.
\newblock Hidden voice commands.
\newblock In {\em USENIX Security Symposium},  513--530.

\bibitem[\protect\citeauthoryear{Carter \bgroup et al\mbox.\egroup
  }{2019}]{carter2019activation}
Carter, S.; Armstrong, Z.; Schubert, L.; Johnson, I.; and Olah, C.
\newblock 2019.
\newblock Activation atlas.
\newblock {\em Distill}.
\newblock https://distill.pub/2019/activation-atlas.

\bibitem[\protect\citeauthoryear{Chattopadhay \bgroup et al\mbox.\egroup
  }{2018}]{chattopadhay2018grad}
Chattopadhay, A.; Sarkar, A.; Howlader, P.; and Balasubramanian, V.~N.
\newblock 2018.
\newblock Grad-cam++: Generalized gradient-based visual explanations for deep
  convolutional networks.
\newblock In {\em 2018 IEEE Winter Conference on Applications of Computer
  Vision (WACV)},  839--847.
\newblock IEEE.

\bibitem[\protect\citeauthoryear{Chen \bgroup et al\mbox.\egroup
  }{2017}]{chen2017ead}
Chen, P.-Y.; Sharma, Y.; Zhang, H.; Yi, J.; and Hsieh, C.-J.
\newblock 2017.
\newblock Ead: elastic-net attacks to deep neural networks via adversarial
  examples.
\newblock {\em arXiv preprint arXiv:1709.04114}.

\bibitem[\protect\citeauthoryear{Dong \bgroup et al\mbox.\egroup
  }{2017}]{dong2017towards}
Dong, Y.; Su, H.; Zhu, J.; and Bao, F.
\newblock 2017.
\newblock Towards interpretable deep neural networks by leveraging adversarial
  examples.
\newblock {\em arXiv preprint arXiv:1708.05493}.

\bibitem[\protect\citeauthoryear{Finlay, Oberman, and
  Abbasi}{2018}]{finlay2018improved}
Finlay, C.; Oberman, A.; and Abbasi, B.
\newblock 2018.
\newblock Improved robustness to adversarial examples using lipschitz
  regularization of the loss.
\newblock {\em arXiv preprint arXiv:1810.00953}.

\bibitem[\protect\citeauthoryear{Goodfellow, Shlens, and
  Szegedy}{2015}]{goodfellow2015explaining}
Goodfellow, I.; Shlens, J.; and Szegedy, C.
\newblock 2015.
\newblock Explaining and harnessing adversarial examples.
\newblock {\em 2015 ICLR} arXiv preprint arXiv:1412.6572.

\bibitem[\protect\citeauthoryear{He \bgroup et al\mbox.\egroup
  }{2016}]{he2016deep}
He, K.; Zhang, X.; Ren, S.; and Sun, J.
\newblock 2016.
\newblock Deep residual learning for image recognition.
\newblock In {\em Proceedings of the IEEE conference on computer vision and
  pattern recognition},  770--778.

\bibitem[\protect\citeauthoryear{Karmon, Zoran, and
  Goldberg}{2018}]{karmon2018lavan}
Karmon, D.; Zoran, D.; and Goldberg, Y.
\newblock 2018.
\newblock Lavan: Localized and visible adversarial noise.
\newblock {\em arXiv preprint arXiv:1801.02608}.

\bibitem[\protect\citeauthoryear{Kurakin, Goodfellow, and
  Bengio}{2016}]{kurakin2016adversarial}
Kurakin, A.; Goodfellow, I.; and Bengio, S.
\newblock 2016.
\newblock Adversarial examples in the physical world.
\newblock {\em arXiv preprint arXiv:1607.02533}.

\bibitem[\protect\citeauthoryear{Kurakin, Goodfellow, and
  Bengio}{2017}]{KurakinGB2016adversarial}
Kurakin, A.; Goodfellow, I.~J.; and Bengio, S.
\newblock 2017.
\newblock Adversarial machine learning at scale.
\newblock {\em 2017 ICLR} arXiv preprint arXiv:1611.01236.

\bibitem[\protect\citeauthoryear{Luo \bgroup et al\mbox.\egroup
  }{2018}]{luo2018random}
Luo, T.; Cai, T.; Zhang, M.; Chen, S.; and Wang, L.
\newblock 2018.
\newblock Random mask: Towards robust convolutional neural networks.

\bibitem[\protect\citeauthoryear{Madry \bgroup et al\mbox.\egroup
  }{2017}]{madry2017towards}
Madry, A.; Makelov, A.; Schmidt, L.; Tsipras, D.; and Vladu, A.
\newblock 2017.
\newblock Towards deep learning models resistant to adversarial attacks.
\newblock {\em arXiv preprint arXiv:1706.06083}.

\bibitem[\protect\citeauthoryear{Papernot \bgroup et al\mbox.\egroup
  }{2016a}]{papernot2016limitations}
Papernot, N.; McDaniel, P.; Jha, S.; Fredrikson, M.; Celik, Z.~B.; and Swami,
  A.
\newblock 2016a.
\newblock The limitations of deep learning in adversarial settings.
\newblock In {\em Security and Privacy (EuroS\&P), 2016 IEEE European Symposium
  on},  372--387.
\newblock IEEE.

\bibitem[\protect\citeauthoryear{Papernot \bgroup et al\mbox.\egroup
  }{2016b}]{papernot2016distillation}
Papernot, N.; McDaniel, P.; Wu, X.; Jha, S.; and Swami, A.
\newblock 2016b.
\newblock Distillation as a defense to adversarial perturbations against deep
  neural networks.
\newblock In {\em Security and Privacy (SP), 2016 IEEE Symposium on},
  582--597.
\newblock IEEE.

\bibitem[\protect\citeauthoryear{Petsiuk, Das, and
  Saenko}{2018}]{petsiuk2018rise}
Petsiuk, V.; Das, A.; and Saenko, K.
\newblock 2018.
\newblock Rise: Randomized input sampling for explanation of black-box models.
\newblock {\em arXiv preprint arXiv:1806.07421}.

\bibitem[\protect\citeauthoryear{Selvaraju \bgroup et al\mbox.\egroup
  }{2017}]{selvaraju2017grad}
Selvaraju, R.~R.; Cogswell, M.; Das, A.; Vedantam, R.; Parikh, D.; and Batra,
  D.
\newblock 2017.
\newblock Grad-cam: Visual explanations from deep networks via gradient-based
  localization.
\newblock In {\em Proceedings of the IEEE International Conference on Computer
  Vision},  618--626.

\bibitem[\protect\citeauthoryear{Sinha, Namkoong, and
  Duchi}{2018}]{sinha2018certifying}
Sinha, A.; Namkoong, H.; and Duchi, J.
\newblock 2018.
\newblock Certifying some distributional robustness with principled adversarial
  training.

\bibitem[\protect\citeauthoryear{Szegedy \bgroup et al\mbox.\egroup
  }{2016}]{Szegedy2016RethinkingTI}
Szegedy, C.; Vanhoucke, V.; Ioffe, S.; Shlens, J.; and Wojna, Z.
\newblock 2016.
\newblock Rethinking the inception architecture for computer vision.
\newblock {\em 2016 IEEE Conference on Computer Vision and Pattern Recognition
  (CVPR)}  2818--2826.

\bibitem[\protect\citeauthoryear{Xiao \bgroup et al\mbox.\egroup
  }{2018}]{xiao2018spatially}
Xiao, C.; Zhu, J.-Y.; Li, B.; He, W.; Liu, M.; and Song, D.
\newblock 2018.
\newblock Spatially transformed adversarial examples.
\newblock In {\em International Conference on Learning Representations}.

\bibitem[\protect\citeauthoryear{Xu \bgroup et al\mbox.\egroup
  }{2019a}]{xu2019topology}
Xu, K.; Chen, H.; Liu, S.; Chen, P.-Y.; Weng, T.-W.; Hong, M.; and Lin, X.
\newblock 2019a.
\newblock Topology attack and defense for graph neural networks: An
  optimization perspective.
\newblock In {\em International Joint Conference on Artificial Intelligence
  (IJCAI)}.

\bibitem[\protect\citeauthoryear{Xu \bgroup et al\mbox.\egroup
  }{2019b}]{xu2018structured}
Xu, K.; Liu, S.; Zhao, P.; Chen, P.-Y.; Zhang, H.; Fan, Q.; Erdogmus, D.; Wang,
  Y.; and Lin, X.
\newblock 2019b.
\newblock Structured adversarial attack: Towards general implementation and
  better interpretability.
\newblock In {\em International Conference on Learning Representations}.

\bibitem[\protect\citeauthoryear{Yang \bgroup et al\mbox.\egroup
  }{2019}]{yang2019ml}
Yang, P.; Chen, J.; Hsieh, C.-J.; Wang, J.-L.; and Jordan, M.~I.
\newblock 2019.
\newblock Ml-loo: Detecting adversarial examples with feature attribution.
\newblock {\em arXiv preprint arXiv:1906.03499}.

\bibitem[\protect\citeauthoryear{Ye \bgroup et al\mbox.\egroup
  }{2019}]{ye2019second}
Ye, S.; Xu, K.; Liu, S.; Cheng, H.; Lambrechts, J.-H.; Zhang, H.; Zhou, A.; Ma,
  K.; Wang, Y.; and Lin, X.
\newblock 2019.
\newblock Adversarial robustness vs model compression, or both?
\newblock {\em International Conference on Computer Vision (ICCV-2019)}.

\bibitem[\protect\citeauthoryear{Yu, Dong, and Chen}{2018}]{yu2018asp}
Yu, F.; Dong, Q.; and Chen, X.
\newblock 2018.
\newblock Asp: A fast adversarial attack example generation framework based on
  adversarial saliency prediction.
\newblock {\em arXiv preprint arXiv:1802.05763}.

\bibitem[\protect\citeauthoryear{Yuan and Lin}{2006}]{yuan2006model}
Yuan, M., and Lin, Y.
\newblock 2006.
\newblock Model selection and estimation in regression with grouped variables.
\newblock {\em Journal of the Royal Statistical Society: Series B (Statistical
  Methodology)} 68(1):49--67.

\bibitem[\protect\citeauthoryear{Zeiler and
  Fergus}{2014}]{zeiler2014visualizing}
Zeiler, M.~D., and Fergus, R.
\newblock 2014.
\newblock Visualizing and understanding convolutional networks.
\newblock In {\em European conference on computer vision},  818--833.
\newblock Springer.

\bibitem[\protect\citeauthoryear{Zhang \bgroup et al\mbox.\egroup
  }{2018}]{zhang2018interpretable}
Zhang, X.; Wang, N.; Ji, S.; Shen, H.; and Wang, T.
\newblock 2018.
\newblock Interpretable deep learning under fire.
\newblock {\em arXiv preprint arXiv:1812.00891}.

\bibitem[\protect\citeauthoryear{Zhou \bgroup et al\mbox.\egroup
  }{2016}]{zhou2016learning}
Zhou, B.; Khosla, A.; Lapedriza, A.; Oliva, A.; and Torralba, A.
\newblock 2016.
\newblock Learning deep features for discriminative localization.
\newblock In {\em Proceedings of the IEEE Conference on Computer Vision and
  Pattern Recognition},  2921--2929.

\end{thebibliography}
\bibliographystyle{aaai}
}

\newpage
\clearpage


\section*{Appendices of \textit{Interpreting Adversarial Examples by Activation Promotion and Suppression}}

\setcounter{section}{0}
\setcounter{figure}{0}
\makeatletter 
\renewcommand{\thefigure}{A\@arabic\c@figure}
\makeatother
\setcounter{table}{0}
\renewcommand{\thetable}{A\arabic{table}}

\section{Attack Generation}\label{app: attack_generation}
\textcolor{black}{
\textbf{IFGSM attack} \citep{goodfellow2015explaining,KurakinGB2016adversarial} produces adversarial examples by performing  iterative fast gradient   sign method  (IFGSM), followed by an $\epsilon$-ball clipping. IFGSM attacks
are designed to be fast, rather than optimal in terms of minimum perturbation.
}

\textcolor{black}{
\textbf{C\&W} \citep{carlini2017adversarial}, \textbf{EAD} \citep{chen2017ead}, and \textbf{\textcolor{black}{Str-} attacks} \citep{xu2018structured} can be unified in the following optimization framework,
{\small\begin{align}\label{eq: attack_general}
\begin{array}{ll}
    \displaystyle \minimize_{\boldsymbol \delta } & f(\mathbf x_0 + \boldsymbol \delta, t) + {\lambda} g(\boldsymbol{\delta})
    \\
  \st    &  (\mathbf x_0 + \boldsymbol \delta) \in [0,1]^n, ~ h(\boldsymbol{\delta}) \leq 0,
\end{array}
\end{align}}%
where $f(\mathbf x_0 + \boldsymbol \delta, t)$ denotes a loss function for targeted misclassification,  $g(\boldsymbol{\delta}) $ is a regularization function that penalizes the norm of adversarial perturbations, $\lambda>0$ is a regularization parameter, and 
$h(\boldsymbol{\delta})$ places optionally hard constraints on $\boldsymbol{\delta}$.
All C\&W, EAD and \textcolor{black}{Str-} attacks enjoy a similar loss function
{\small \begin{align} \label{eq: fx_logit}
f(\mathbf x_0 + \boldsymbol \delta, t) =  c \cdot \max \{ & \max_{j \neq t}  Z(\mathbf x_0+ \boldsymbol \delta)_j \nonumber\\ & -Z(\mathbf x_0+ \boldsymbol \delta )_t, - \kappa \},
\end{align}}%
where $Z(\mathbf x)_j$ is the $j$-th element of logits $Z(\mathbf x)$, namely, the output before the last softmax layer in CNNs, and $\kappa$ is a confidence parameter. Clearly, as $\kappa$ increases, the minimization of $f$ would reach the target label with high confidence. In this paper, we set  $\kappa = 1$ by default. \textcolor{black}{It is worth mentioning that problem \eqref{eq: attack_general} can be efficiently solved via alternating direction method of multipliers (ADMM) \citep{boyd2011distributed,xu2018structured}, regardless of whether or not $g(\bm \delta)$ is differentiable.}
}

\textcolor{black}{
\textbf{C\&W  attack} \citep{carlini2017adversarial} adopts the $\ell_p$ norm to penalize the strength of adversarial perturbations $\boldsymbol{\delta}$, namely, $g(\boldsymbol{\delta}) = \| \boldsymbol{\delta }\|_{p}$ and $h(\boldsymbol{\delta}) = 0$ in \eqref{eq: attack_general}, where $p \in \{0,2,\infty \}$. In practice, the squared $\ell_2$ norm is commonly used.
}

\textcolor{black}{
\textbf{EAD attack} \citep{chen2017ead} specifies the regularization term
$\lambda g(\boldsymbol{\delta})$ as an  elastic-net regularizer $ \lambda_1 \|\boldsymbol \delta  \|_2^2 + \lambda_2 \|\boldsymbol \delta  \|_1$ in \eqref{eq: attack_general}, and  $h(\boldsymbol{\delta}) = 0$. 
It has empirically shown that
the use of  elastic-net regularizer improves the   transferability of adversarial examples. 
}

\textcolor{black}
{\textbf{
\textcolor{black}{Str-}attack} \citep{xu2018structured} takes into account the 
group-level sparsity of adversarial perturbations by choosing $g(\boldsymbol{\delta})$ as the group Lasso penalty \citep{yuan2006model}. In the mean time, it constrains the pixel-level perturbation by setting $h(\boldsymbol \delta) = \| \boldsymbol{\delta }\|_\infty - \epsilon$ for a  tolerance $\epsilon >0$. }

\section{CAM-based Interpretation}
\label{app: adv_example_cam}

\begin{figure}[h]
\centerline{
\begin{tabular}{cc}
      \includegraphics[width=0.23\textwidth]{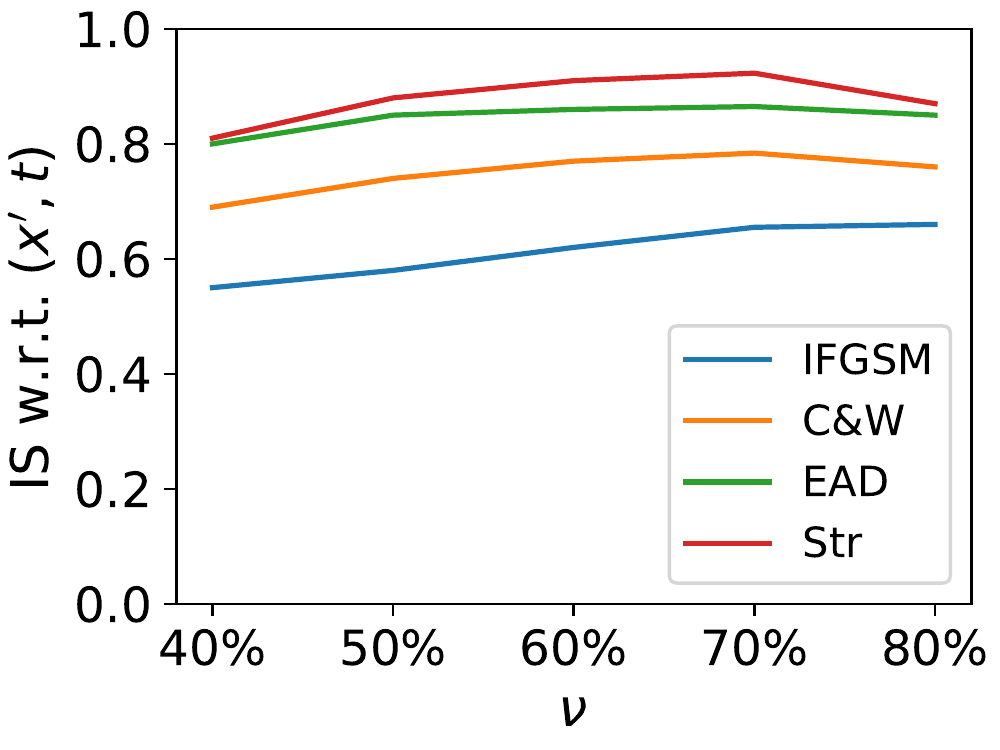} 
      &\includegraphics[width=0.23\textwidth]{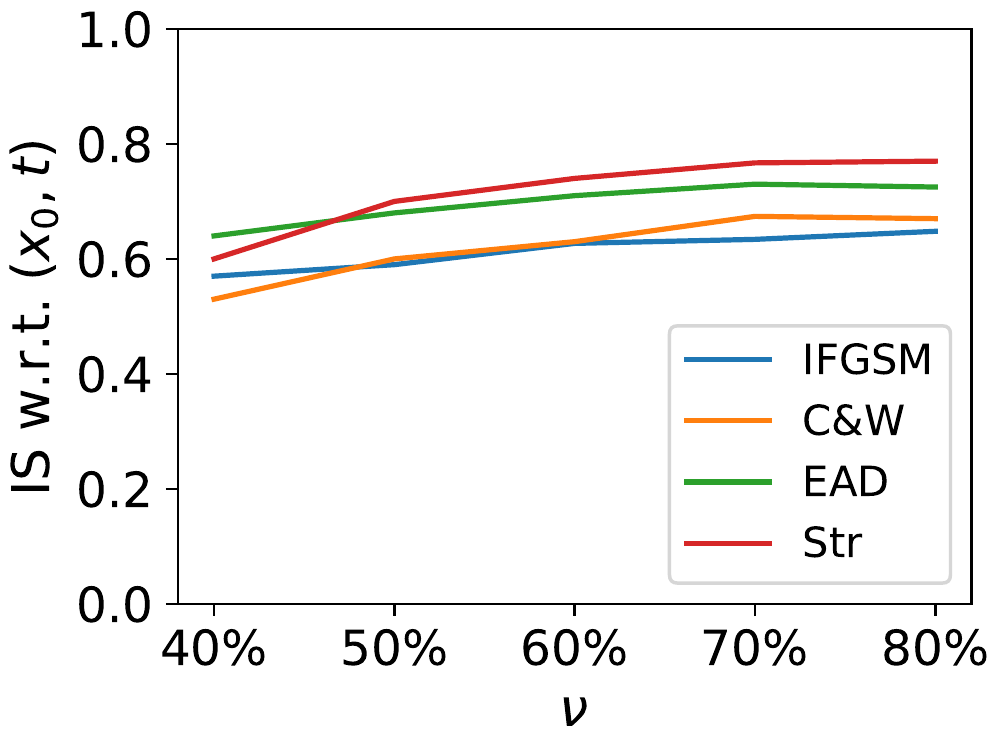} 
\end{tabular}}
\vspace{-3mm}
\caption {\footnotesize{IS versus $\nu$ for $4$  attack types.
}}
\label{fig: v_check}
\end{figure}

\textbf{Sensitivity of hyper-parameter $\nu$ to IS.} Figure\,\ref{fig: v_check} presents IS against $\nu$ for IFGSM, C\&W, EAD and Str attacks.

\begin{figure}[htb]
\centering{
\begin{tabular}{c}
\includegraphics[width=.45\textwidth,height=!]{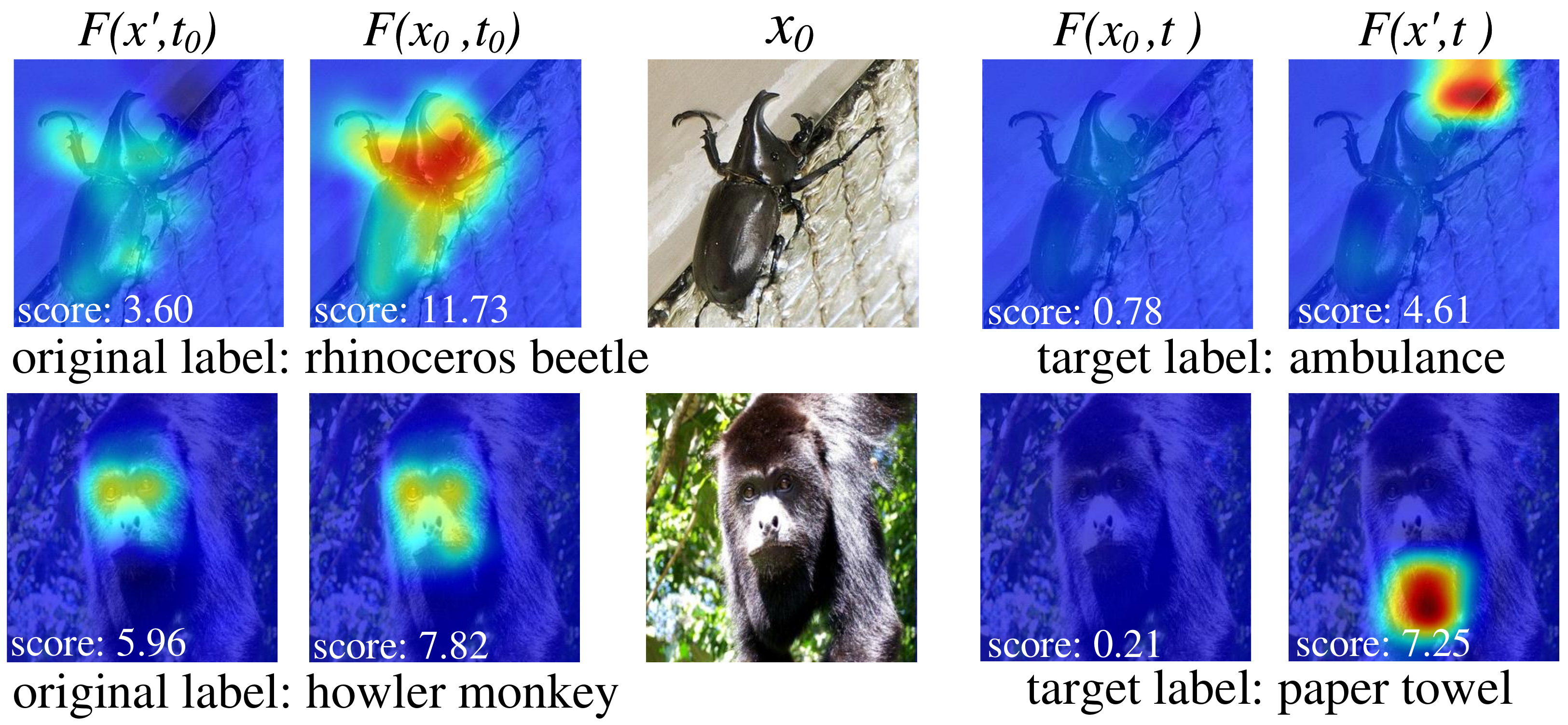}  
\end{tabular}}
\caption{\footnotesize{CAMs of two natural/adversarial examples (in rows), generated by C\&W attack, where
$F(\mathbf x^\prime, t_0)$, $F(\mathbf x_0, t_0)$, $\mathbf x_0$, $F(\mathbf x_0, t)$, $F(\mathbf x^\prime, t)$ are shown from the left to the right at each row. 
}}
\label{fig: CAM_example}
\end{figure}

\begin{figure}[!b]  
  \centering
\vspace*{-10mm}  
\hspace*{-0.07in}\includegraphics[width=0.45\textwidth]{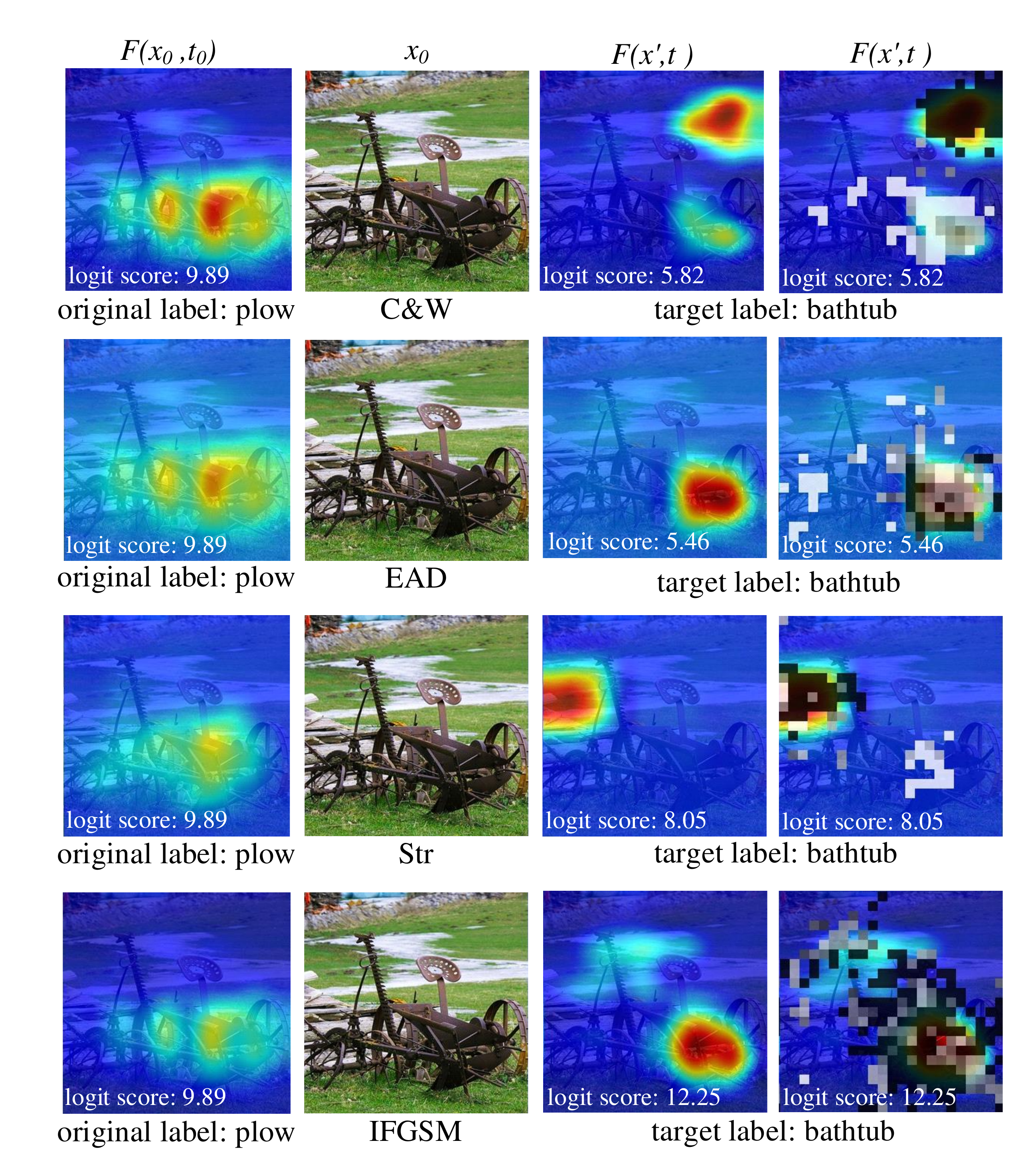}
\caption{\footnotesize{Four adversarial examples with CAM visualization generated by C\&W, EAD, Str, and IFGSM attacks, respectively.
Left to right: $F(\mathbf x_0, t_0)$, $x_0$, $F(\mathbf x^\prime, t)$, and overlaid PSR $r_i$ on $F(\mathbf x^\prime, t)$ at locations of the top $70\%$ most significant   perturbed grids {ranked by $s_i$}. Here CAMs at each row are normalized with respect to their maximum value.}}
\label{fig: CAM_example_compare}
\end{figure}

\begin{figure}[b]
   \centering
\begin{tabular}{p{0.01in}p{0.85in}p{0.01in}p{0.85in}p{0.85in}}
&  \hspace*{-0.05in}  \parbox{0.75in}{\centering \footnotesize original } 
&  
& \hspace*{-0.05in} \parbox{0.75in}{\centering \footnotesize  C\&W attack } 
& \hspace*{-0.2in} \parbox{0.75in}{\centering \footnotesize  Str-attack}
\\
&    
\includegraphics[width=0.75in]{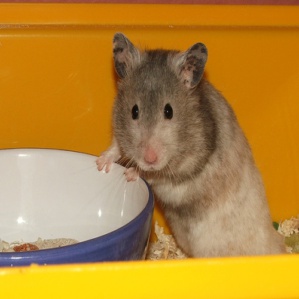}&  \hspace*{-0.12in} 
\rotatebox{90}{\parbox{0.75in}{\centering \footnotesize  PSR over perturbed grids}} & 
\includegraphics[width=0.75in]{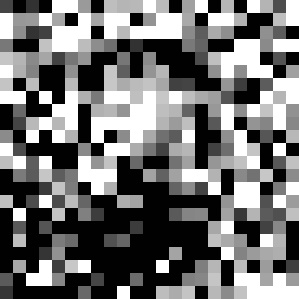}& \hspace*{-0.2in} 
\includegraphics[width=0.75in]{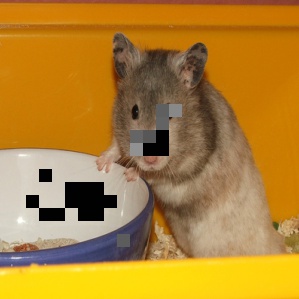}
\\
\hspace*{0.03in}\rotatebox{90}{\parbox{0.75in}{\centering \footnotesize  CAM w.r.t. $t_0$}}  &  
\hspace*{-0.03in}
\includegraphics[width=0.75in]{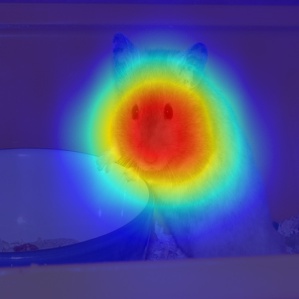}&  
&  
\includegraphics[width=0.75in]{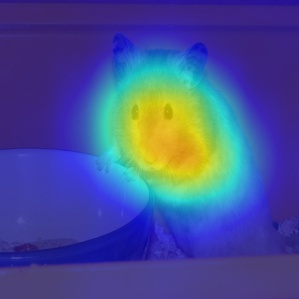}&
\hspace*{-0.2in} 
\includegraphics[width=0.75in]{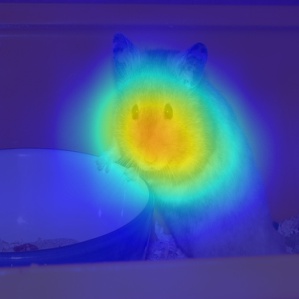}
 \\
\hspace*{0.03in} \rotatebox{90}{\parbox{0.75in}{\centering \footnotesize    CAM w.r.t. $t$ }} & \hspace*{-0.03in}
\includegraphics[width=0.75in]{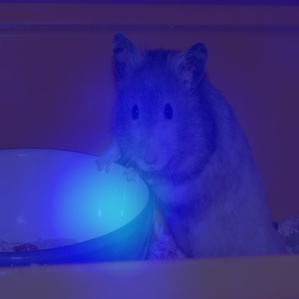}&  
& 
\includegraphics[width=0.75in]{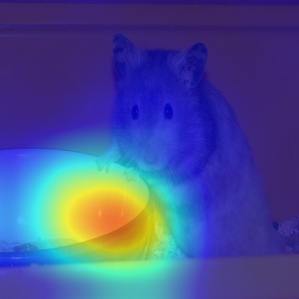}&\hspace*{-0.2in} 
\includegraphics[width=0.75in]{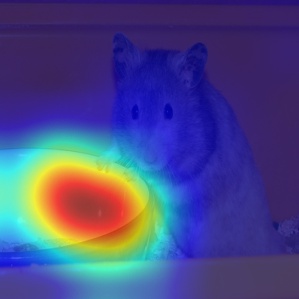}
\end{tabular}
\caption{\footnotesize{Visual explanation of the `hamster'-to-`cup' example crafted by   C\&W   and   Str-attack, where the true label $t_0$ is  `hamster', and the target label $t$ is `cup'. The first row is the natural image and PSRs over perturbed grids. The second (third) row is CAM with respect to the column-wise natural/adversarial example and the row-wise label.
}}
    \label{fig: multipleobjects}
\end{figure}

\begin{figure*}[htb]
\centerline{
\begin{tabular}{cc}
       \includegraphics[width=.45\textwidth]{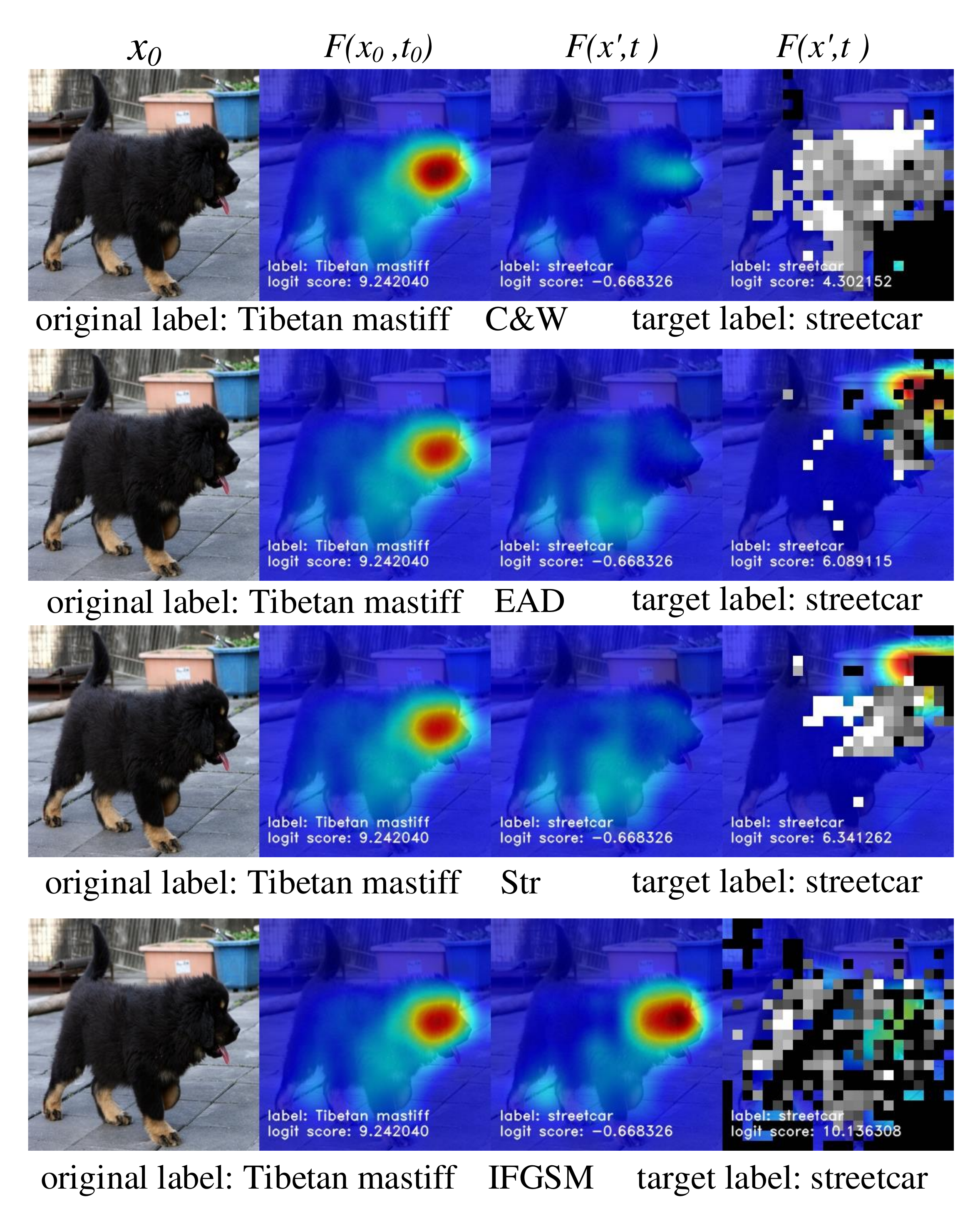}
    &  \includegraphics[width=.45\textwidth]{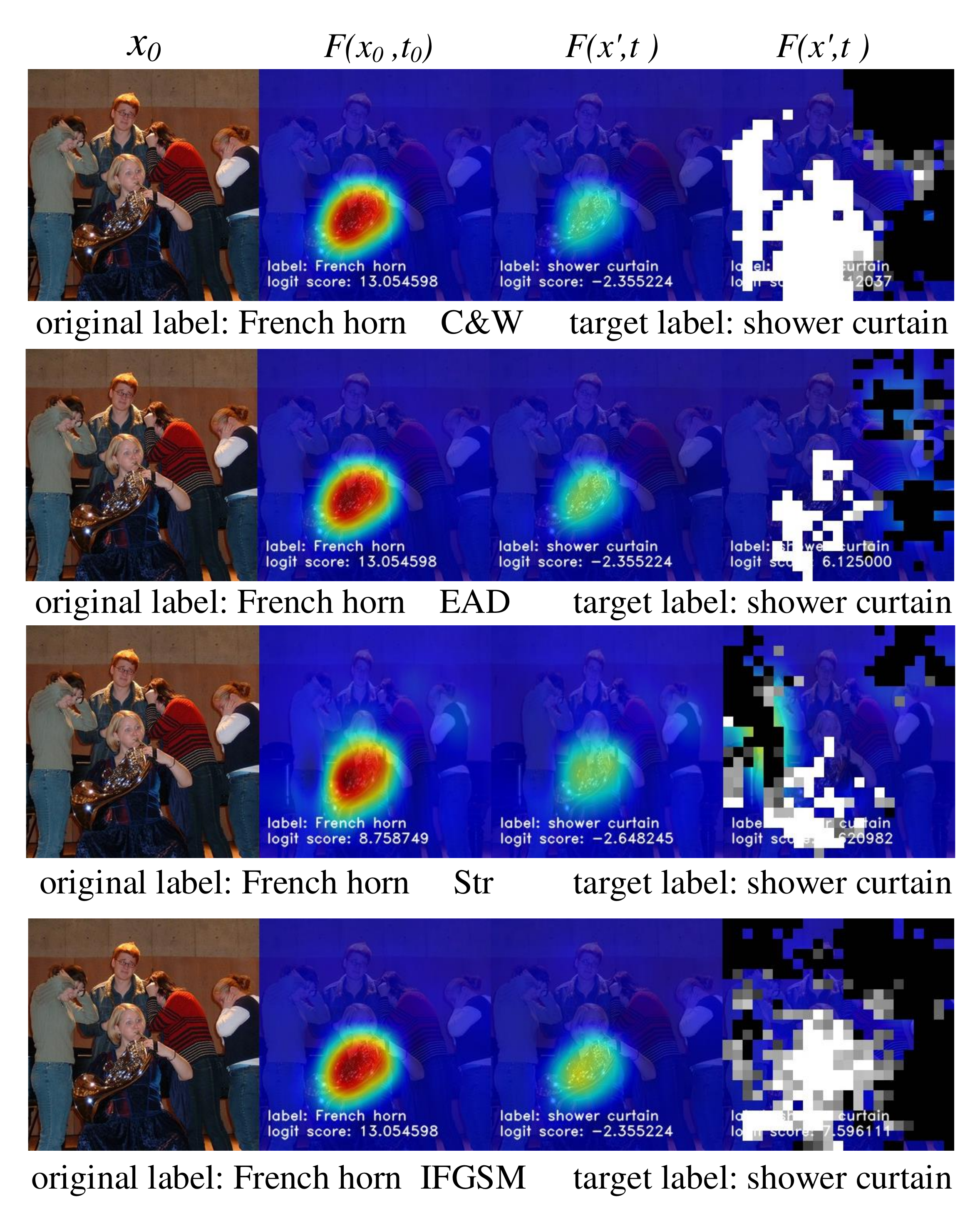}   
\end{tabular}}
\caption{\footnotesize{ Attacking images with complex background under C\&W, EAD, \textcolor{black}{Str-}, and IFGSM attacks.
}}
  \label{fig: supp1}
\end{figure*}

\begin{figure}[htb]
   \centering
\begin{tabular}{p{0.48in}p{0.001in}p{0.42in}p{0.42in}p{0.42in}p{0.42in}}
  \makecell{\footnotesize \  original }&
 & \makecell{ \footnotesize \ \  C\&W } 
 & \makecell{ \footnotesize \ \ EAD  }
 & \makecell{\footnotesize  \ Str } 
 & \makecell{ \footnotesize \  IFGSM }  
 \\
\includegraphics[width=0.56in]{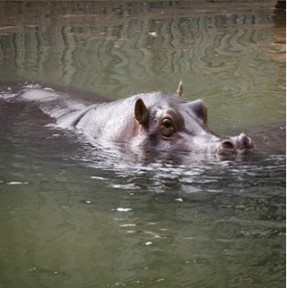}&  
\rotatebox{90}{ \footnotesize w/o refine} &    \hspace*{-0.06in} 
\includegraphics[width=0.56in]{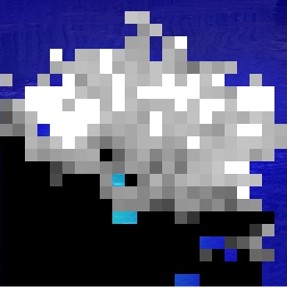}& \hspace*{-0.06in} 
\includegraphics[width=0.56in]{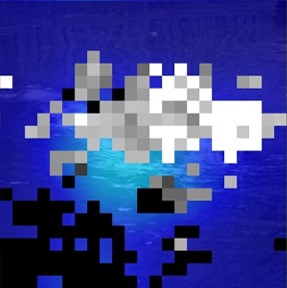}&\hspace*{-0.06in} 
\includegraphics[width=0.56in]{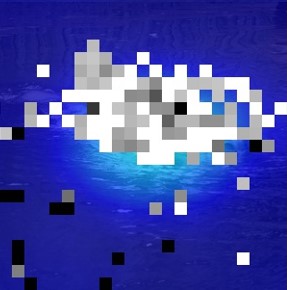}& \hspace*{-0.06in} 
\includegraphics[width=0.565in]{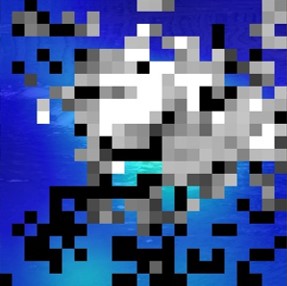}
 \\
\includegraphics[width=0.56in]{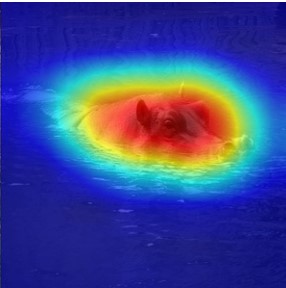}& 
\rotatebox{90}{ \footnotesize with refine  } &    \hspace*{-0.06in} 
\includegraphics[width=0.56in]{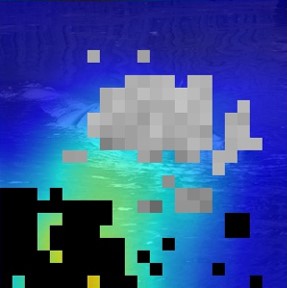}&   \hspace*{-0.06in} 
\includegraphics[width=0.56in]{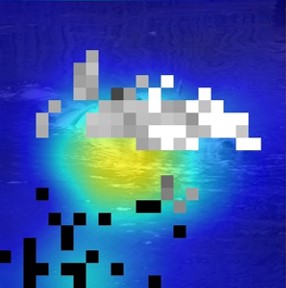}&   \hspace*{-0.06in} 
\includegraphics[width=0.565in]{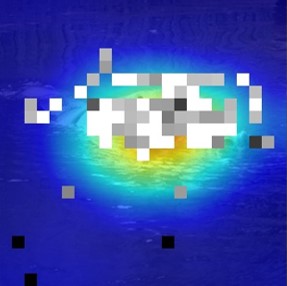}&   \hspace*{-0.06in} 
\includegraphics[width=0.56in]{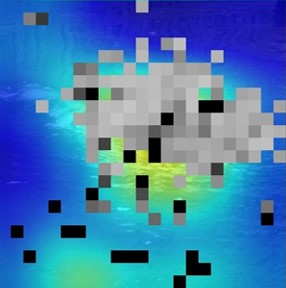}
\end{tabular}
\caption{\footnotesize{\textcolor{black}{
The
`hippopotamus'-to-`streetcar' adversarial example with and without refinement under $\mathcal S_1$. Here the left-bottom subplot shows CAM of the original image w.r.t. the true label `hippopotamus', and the right subplots present
PSRs of unrefined and refined grid-level perturbations   overlaid on CAMs of adversarial examples w.r.t. the target label `streetcar'.}
}}
    \label{fig: refine1}
\end{figure}

\noindent \textbf{Additional results on CAM-based interpretation.}
In Figure\,\ref{fig: CAM_example}, we demonstrate more examples of class-specific discriminative regions visualized by CAM, namely, $F(\mathbf x^\prime, t_0)$, $F(\mathbf x_0, t_0)$, $\mathbf x_0$, $F(\mathbf x_0, t)$, $F(\mathbf x^\prime, t)$. 
In Figure\,\ref{fig: CAM_example_compare}, we fix the orginal image together with its true and target labels to visualize the difference of attack methods through CAM.
In Figure\,\ref{fig: onevsone}, we present the adversarial attack of multiple images with a fixed source-target label pair. As we can see, the balance-dominated perturbation pattern appears at the discriminative region of `eagle'.
In Figure\,\ref{fig: multipleobjects}, we present
  the `hamster'-to-`cup' example, where objects of the  original label and   the target label exist   
simultaneously. We observe that the adversary shows  suppression
on the discriminative region of the original label and promotion  on the discriminative region of  the target label.
Compared to the C\&W attack,  Str-attack is more effective in both suppressing and promotion since it   perturbs less pixels. 
 \textcolor{black}{In Figure\,\ref{fig: supp1}},   images involve   more
heterogeneous and complex backgrounds.  As we can see, an effective adversarial attack (e.g., Str-attack)  
perturbs less but more meaningful pixels, which have a
better correspondence with the discriminative image regions of the original and target classes. 
 In Figure\,\ref{fig: refine1}, we present a
 `hippocampus'-to-`streetcar' example with refined attacks under $\mathcal S_1$. As we can see, 
 it is possible to obtain a more effective attack by perturbing less but `right'   pixels (i.e., with better correspondence with discriminative image regions).

\section{Effectiveness of Refined Adversarial Pattern}\label{app: attack_refine}
We consider the following    unified optimization problem to refine adversarial attacks  
{\small    \begin{align}\label{eq: attack_general_regin}
\begin{array}{ll}
    \displaystyle \minimize_{\boldsymbol \delta } & f(\mathbf x_0 + \boldsymbol \delta, t) + \lambda g(\boldsymbol{\delta})
    \\
  \st    &  (\mathbf x_0 + \boldsymbol \delta) \in [0,1]^n, ~ h(\boldsymbol{\delta}) \leq 0 \\
  &  \delta_i = 0, ~\text{if}~ i \notin \mathcal S_k,~ \text{$k=1$ or $2$},
\end{array}
\end{align}}%
where we represent  the  refinement operations  (a) and (b) through the constraint sets
$\mathcal S_1 = \{ \forall i \, | \, s_i > \beta  \}$ for a  positive threshold $\beta$
\footnote{\textcolor{black}{We sort  $\{ s_i\}$ to $\{ \tilde{s}_i\}$ in an ascending order, and  set $\beta = \tilde s_k$ for the smallest $k$ with $\sum_{i=1}^k \tilde{s_i} /\sum_{i=1}^m \tilde{s_i} \geq 30\%$. We filter less significant perturbations under their cumulative power.}}  and $\mathcal S_2 = \{ \forall i \, | \, [B(\mathbf x_0, t_0)]_i > 0 \}$.  
In $\mathcal S_1$,
 $s_i$ defined by \eqref{eq: s_d} characterizes the strength of the adversarial pattern.  In $\mathcal S_2$, $B(\mathbf x_0, t_0)$ defined by \eqref{eq: B} localizes pixels corresponding to the   most  discriminative  region  associated with the true label. 
 \textcolor{black}{Problem \eqref{eq: attack_general_regin} can similarly be    solved as \eqref{eq: attack_general}, with an additional projection on the sparse constraints given by $\mathcal S_1$ and $\mathcal S_2$.}

  We present the   effectiveness of attacks  with refinement  under $\mathcal S_1$ in  
Table\,\ref{table:refine}. Here the effectiveness of an attack   is characterized by  
its attack success rate (ASR) as well as  $\ell_p$-norm distortions. We find that many pixel-level adversarial perturbations are redundant, in terms of the reduction in the $\ell_0$ norm\footnote{$\| \mathbf x \|_0$: $\#$ of nonzero elements in $\mathbf x$.} of $\boldsymbol{\delta}$, which can be removed   without losing   effectiveness in the attack success rate and $\ell_p$-norm distortions for $p > 0$.
 
 \section{Interpretation via Network Dissection}\label{app: dissection}
\begin{figure}[htb]
   \centering
\begin{tabular}{ccc}

\footnotesize{adv. examples} & 
 \begin{tabular}[c]{@{}c@{}}\footnotesize{$F(\bm x_0 , t_0)$} \end{tabular} 
  &\begin{tabular}[c]{@{}c@{}}\footnotesize{ $F(\bm x^\prime , t)$} \& \footnotesize{PSRs}  \end{tabular}
 \\[0pt]

\includegraphics[width=0.95in]{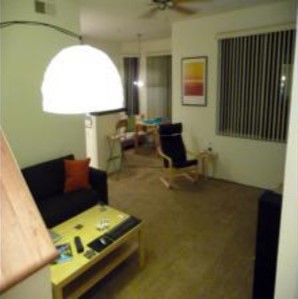}&
\includegraphics[width=0.95in]{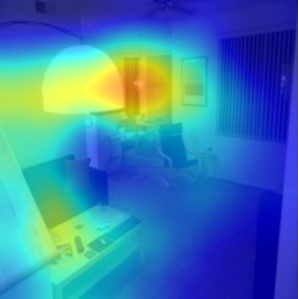}&
\includegraphics[width=0.95in]{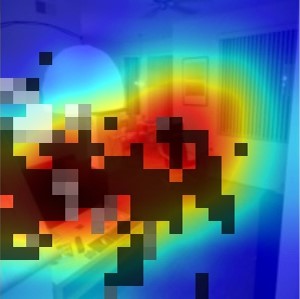} 
\\[-2pt]

\includegraphics[width=0.95in]{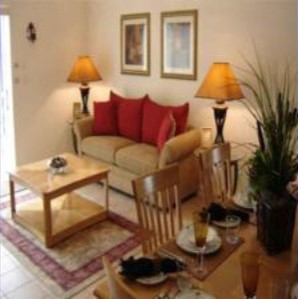}&
\includegraphics[width=0.95in]{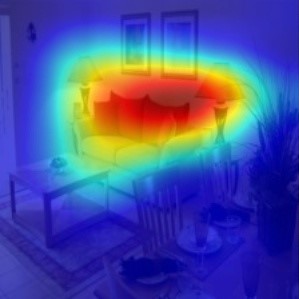}&
\includegraphics[width=0.95in]{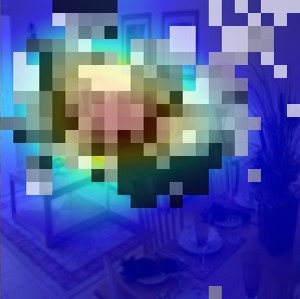} \\[-3pt]
 \multicolumn{3}{l}{\footnotesize{ $t_0$: table lamp --  $t$: studio coach} }
\\[6pt]

\includegraphics[width=0.95in]{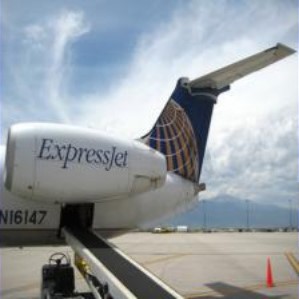}&
\includegraphics[width=0.95in]{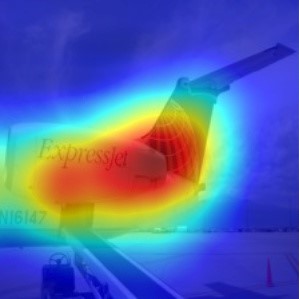}&
\includegraphics[width=0.95in]{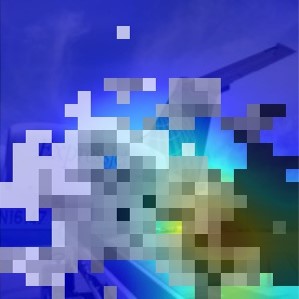} \\[-3pt]

\includegraphics[width=0.95in]{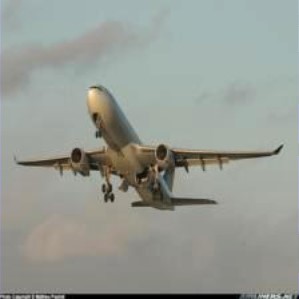}&
\includegraphics[width=0.95in]{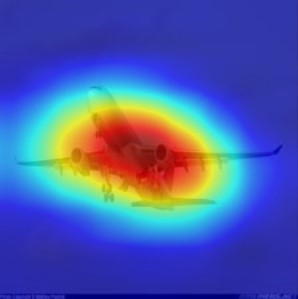}&
\includegraphics[width=0.95in]{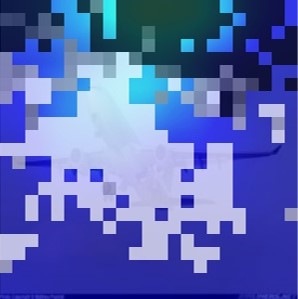} \\[-3pt]
 \multicolumn{3}{l}{\footnotesize{ $t_0$: airliner --  $t$: seashore} }
\\

\end{tabular}
\caption{\footnotesize{Interpreting adversarial perturbations via CAM and PSR.  Image examples are from   Figure \,\ref{fig: dissec}. For PSR, only the top $70\%$ most significant   perturbed grids {ranked by   $\{s_i\}$ \eqref{eq: s_d}} are shown. The white and black colors represent
the suppression-dominated regions   ($r_i < -1$) and the promotion-dominated regions  ($r_i > 1$), respectively. The gray color  corresponds to balance-dominated perturbations  ($r_i \in [-1,1]$).
}
}
    \label{fig: CAM_Netdisecction}
\end{figure}

In Figure\,\ref{fig: CAM_Netdisecction}, we connect images in Figure\,\ref{fig: dissec} to PSR and CAM based visual explanation. For example, the suppressed image region identified by PSR (white color) corresponds to the interpretable activation of object concept airplane in Figure\,\ref{fig: dissec}. And the promoted image region identified by PSR (black color) corresponds to the interpretable activation of scene concept beach.

In Table\,\ref{table:correlation}, we report 
 the clean test accuracy (CTA) and 
the adversarial test accuracy (ATA)    
under three masking settings on the last convolutional layer of $\mathrm{conv}5$: a) 
our proposed masking over top 
  $10$  sensitive       units with concept-level interpretability, b) random masking over $10$ units, and c) no masking.
ATA is   obtained by perturbing $1000$ randomly selected   test images using $k$-step PGD attack \citep{madry2017towards}, where $k\in \{ 10, 20, 50, 100\}$.
Our preliminary results show that the proposed approach yields the highest ATA, 
  balanced with  slight degradation on CTA.



\clearpage

\begin{figure}[htb]
\centerline{
\begin{tabular}{c}
      \includegraphics[width=.45\textwidth]{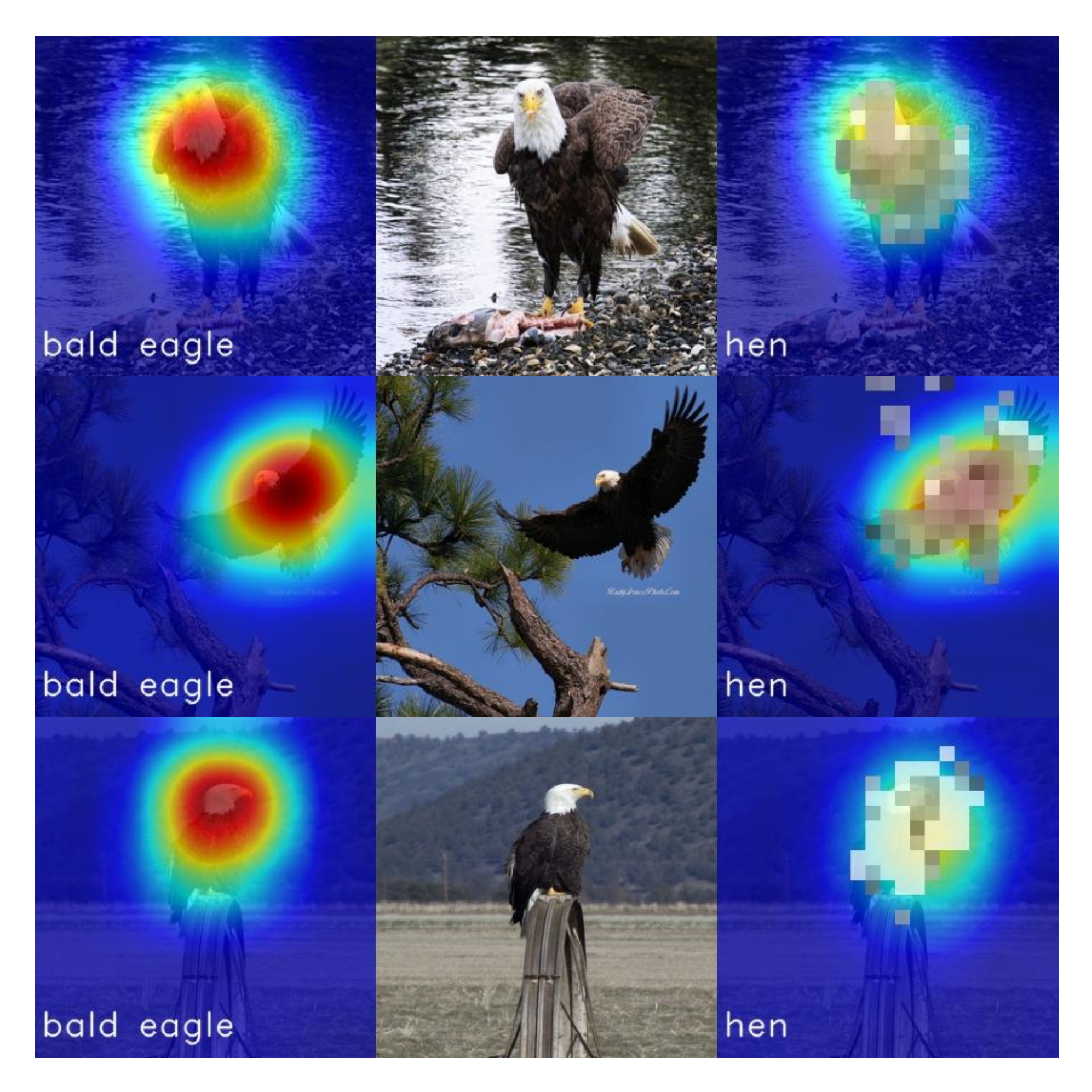}  
\end{tabular}}
\caption {\footnotesize{Multiple images with a fixed source-target label pair: CAM with respect to the source label `bald eagle' (1 column), original image `bald eagle' (2 column), CAM  with respect to target label `hen' together with C\&W 
perturbation patterns (3 column), which is
measured by promotion-suppression ratio (PSR), i.e.  suppression-  (white), promotion-  (black), and balance-dominated adversaries (gray).
}}
  \label{fig: onevsone}
\end{figure}

 \begin{table*}[t]
\centering
\caption{ \textcolor{black}{Attack performance of adversarial perturbations with and without refinement under  $\mathcal S_1$ over $5000$ images.}  
}
\begin{adjustbox}{max width=0.8\textwidth }
\begin{tabular}{c|c|c|c|c|c|c|c|c|c|c}
\hline
\toprule[1pt]
\multirow{2}{*}{attack}                 & \multirow{2}{*}{model}  & \multicolumn{2}{c|}{$\ell_0$} & \multicolumn{2}{c|}{$\ell_1$} & \multicolumn{2}{c|}{$\ell_2$} & \multicolumn{2}{c|}{$\ell_\infty$} & ASR\\ \cline{3-11} 
                       &        & original ($\boldsymbol{\delta}$)   & refine ($ {\boldsymbol{\delta}}_{\mathcal S}$)   & original    & refine   & original    & refine   & original     & refine    & refine\\  \midrule[1pt]
\multirow{2}{*}{IFGSM} & Resnet & 266031      & 61055     & 1122.56     & 176.08    & 2.625       & 1.87      & 0.017        & 0.035      & 96.7\%\\ 
                       & Incep. & 266026      & 59881     & 812.94      & 155.89    & 1.926       & 1.22      & 0.019        & 0.033      &100\%\\  \midrule[1pt]
\multirow{2}{*}{C\&W}  & Resnet & 268117      & 21103     & 183.65      & 134.26    & 0.697       & 0.727     & 0.028        & 0.029      &100\%\\ 
                       & Incep. & 268123      & 22495     & 144.94      & 96.75     & 0.650       & 0.673     & 0.028        & 0.034      &100\%\\ \midrule[1pt]
\multirow{2}{*}{EAD}   & Resnet & 66584       & 20147     & 42.57       & 63.28     & 1.520       & 1.233     & 0.234        & 0.096      &100\%\\ 
                       & Incep. & 69677       & 18855     & 30.17       & 45.88     & 1.289       & 1.107     & 0.229        & 0.083     & 100\%\\  \midrule[1pt]
\multirow{2}{*}{Str}   & Resnet & 30823       & 18744     & 119.76      & 110.54    & 1.250       & 1.132     & 0.105        & 0.087      &100\%\\ 
                       & Incep. & 27873       & 15967     & 86.55       & 82.33     & 1.174       & 0.985     & 0.103        & 0.072      &100\%\\ \bottomrule[1pt]
\end{tabular}
\end{adjustbox}
\label{table:refine}
\end{table*}

\begin{table*}[h] 
\centering
\caption{Evaluation of neuron masking on clean test accuracy (CTA) and adversarial test accuracy (ATA) against $k$-step PGD attacks.}
\label{table:correlation}
 \begin{adjustbox}{max width=0.6\textwidth }
 \vspace*{-2mm}
\begin{tabular}{c|c|c|c|c|c}
\hline
\toprule[1pt]
  &  CTA    
&  
\begin{tabular}[c]{@{}c@{}} ATA \\ ($10$-PGD)   \end{tabular}
& \begin{tabular}[c]{@{}c@{}} ATA \\ ($20$-PGD)   \end{tabular}  & \begin{tabular}[c]{@{}c@{}} ATA \\ ($50$-PGD)   \end{tabular} & \begin{tabular}[c]{@{}c@{}} ATA \\ ($100$-PGD)   \end{tabular}
\\ \midrule[1pt]
ours 
&  75.9\% 
& 60.1\% & 51.5\% & 39.7\% & 33.6\%
\\ \midrule[1pt]
random masking 
& 77.0\% & 
40.5\% & 28.7\% & 18.3\% & 14.3\%
\\ \midrule[1pt]
 no masking
 &  78.2\%  & 
 41.7\% & 31.0\% & 18.8\% & 15.7\%
\\ \midrule[1pt]                      
\end{tabular}
\vspace{-5mm}
\end{adjustbox}
\end{table*}

\end{document}